\documentclass[accepted]{uai2026} 
                        
\usepackage[american]{babel}

\usepackage{natbib} 
\usepackage{mathtools} 
\usepackage{booktabs} 
\usepackage{tikz} 
\usepackage{amsmath}
\usepackage{amssymb}
\usepackage{mathtools}
\usepackage{amsthm}
\usepackage{algorithm}
\usepackage{algorithmic}
\usepackage{graphicx}
\usepackage{placeins}

\title{Unsupervised Continual Learning for Amortized Bayesian Inference}

\author[1]{\href{mailto:<aayush.mishra@tu-dortmund.de>?Subject=Your UAI 2026 paper}{Aayush Mishra}{}}
\author[1]{Šimon Kucharský}
\author[1]{Paul-Christian Bürkner}
\affil[1]{%
    Department of Statistics\\
    TU Dortmund University\\
    Germany
}

\begin{document}
\maketitle

\begin{abstract}
Amortized Bayesian Inference (ABI) enables efficient posterior estimation using generative neural networks trained on simulated data, but often suffers from performance degradation under model misspecification. While self-consistency (SC) training on unlabeled empirical data can enhance network robustness, current approaches are limited to static, single-task settings and fail to handle sequentially arriving data or distribution shifts. We propose a continual learning framework for ABI that decouples simulation-based pre-training from unsupervised sequential SC fine-tuning on real-world data. To address the challenge of catastrophic forgetting, we introduce two adaptation strategies: (1) SC with episodic replay, utilizing a memory buffer of past observations, and (2) SC with elastic weight consolidation, which regularizes updates to preserve task-critical parameters. Across three diverse case studies, our methods significantly mitigate forgetting and yield posterior estimates that outperform standard simulation-based training, achieving estimates closer to MCMC reference, providing a viable path for trustworthy ABI across a range of different tasks.

\end{abstract}

\section{Introduction} \label{sec:intro}

Amortized Bayesian Inference \citep[ABI;][]{radev2023bayesflow, schmitt2023detecting, elsemuller2023sensitivity, zammit2025neural, gloeckler2024all, deistler2025simulation, wildberger2023flow} addresses computational limitations of traditional Bayesian inference methods by training generative neural networks to approximate posterior distributions over parameters $\theta$ using simulated pairs of data $x$ and parameters $\theta$ from a probabilistic model $p(\theta, x)$. Once trained, these networks enable fast, near-instant inference across a large number of observations, effectively amortizing the cost of inference over many queries. 

Despite its efficiency, ABI remains sensitive to model misspecification and domain shifts, which can severely degrade accuracy of neural posterior estimates. Consequently, a growing body of work focuses on improving the robustness of ABI \citep{huang2023learning, wehenkel2024addressing, schmitt2023leveraging, mishra2025robust, elsemuller2025does}. One promising direction is the use of Bayesian self-consistency losses \citep{schmitt2023leveraging, mishra2025robust, kucharsky2025towards} which enforce consistency between the learned posterior approximation and the generative model using unlabeled real-world data.

Up until now, existing robustness approaches including self-consistency losses implicitly assume a static or single-task setting. However, in many real-world applications, data arrive sequentially over time rather than being available all at once, such as galaxy observations from different telescopes  \citep{ghosh2023morphological, tian2025automatic}, gravitational-wave detections \citep{abbott2023gwtc}, and medical imaging across devices \citep{wu2024continual}. In other contexts, the same ABI model may be used for inference across a range of different scientific domains.

This motivates continual learning (CL) approaches, as retraining models from scratch each time new data become available is computationally expensive, often impractical, and diminishes the pay-off of amortization. CL aims to enable models to acquire new knowledge from incoming data while preserving performance on previous data, thereby mitigating the problem of catastrophic forgetting \citep{wang2024comprehensive, van2019three, parisi2019continual, rolnick2019experience}. Referring to the sequential incoming datasets as \emph{tasks}, the central challenge lies in ensuring that learning new tasks does not overwrite the representations or network parameters critical for earlier tasks. CL also enables the network to transfer knowledge from earlier tasks, improving data efficiency when adapting to new tasks through shared representations and replay mechanisms \citep{rolnick2019experience, parisi2019continual}.

This article makes the following contributions:
\begin{itemize}
    \item We propose a two-stage training for ABI that decouples supervised simulation-based training from unsupervised self-consistency training on empirical data, allowing us to fine-tune networks for new incoming tasks in an unsupervised CL setting.
    
    \item We demonstrate a key failure mode of ABI: when applied sequentially across tasks, naive self-consistency updates can cause catastrophic forgetting.

    \item We introduce \textbf{self-consistency with episodic replay}, a CL method that mitigates catastrophic forgetting by retaining a small buffer of data from previous tasks; we further provide a memory-budgeted variant that uses clustering to preserve diversity of training data while keeping memory requirements constant.

    \item We introduce \textbf{self-consistency with elastic weight consolidation}, an unsupervised regularization based approach which prevents catastrophic forgetting by constraining weights important for previous tasks while allowing less critical parameters to adapt flexibly to new tasks.

    \item We further examine a unified objective that combines episodic replay and elastic weight consolidation with self-consistency, enabling us to analyze the combined effects of both CL mechanisms.

    \item Across three case studies, we demonstrate that the proposed methods substantially improve accuracy of posterior approximations on empirical data across different tasks against simulation-based training. Further, we show that our unsupervised continual learning approaches effectively mitigate catastrophic forgetting.
\end{itemize}

\section{Methods} \label{sec:methods}



\subsection{Neural Posterior Estimation}

We consider the general Bayesian inference problem of estimating parameters $\theta$ from data $x$ addressed by means of generative neural networks. The posterior is defined by Bayes’ rule, $p(\theta \mid x) \propto p(\theta, x) = p(x \mid \theta) \, p(\theta)$, where $p(\theta)$ is the prior and $p(x \mid \theta)$ the likelihood. 
ABI casts posterior inference as a supervised learning problem.  It learns a conditional distribution $q_{\Phi}(\theta \mid x)$, via a neural network with weights $\Phi$, to approximate the posterior $p(\theta \mid x)$ by optimizing a simulation-based (SB) loss of the form
\begin{equation}
\mathcal{L}_{\text{SB}}(\Phi)
=
\mathbb{E}_{(\theta, x) \sim p(\theta,x)}
\left[
J\big(q_\Phi(\cdot \mid x), \theta\big)
\right].
\end{equation}
Here, $J$ is a strictly-proper scoring rule such as maximum likelihood loss for discrete normalizing flows \citep{papamakarios2021normalizing, kobyzev2020normalizing} or vector-field loss for flow-matching \citep{lipman2022flow}. As we require fast density estimation for losses used for CL training, we use discrete normalizing flows via invertible neural networks. 
The neural networks to be optimized are a generative network $q$ and (optionally) a summary network $h$ extracting lower dimensional sufficient statistics from the data. These networks are optimized using the maximum-likelihood loss
%
%
\begin{equation}
    \mathcal{L}_{\text{SB}}(\Phi)
    = \mathbb{E}_{p(\theta, x)} \left[ - \log q_{\phi}(\theta \mid h_{\psi}(x)) \right],
    \label{eq:ex-npe}
\end{equation}
%
%
where $\Phi = (\phi, \psi)$ denotes the collection of all trainable parameters of the network. We approximate the expectations in Eq. \ref{eq:ex-npe} with finite simulated data as
\begin{equation}
    \mathcal{L}_{\text{SB}}(\Phi) \approx - \frac{1}{N} \sum_{n=1}^{N} \log q_{\phi}\!\left(\theta_n \mid h_{\psi}(x_n)\right). 
    \label{eq:npe}
\end{equation}
where $(\theta_n, x_n) \sim p(\theta, x)$ are simulations from the joint probabilistic model. 

While simulation-based training performs well when a large amount of simulated training data is available and the underlying model is correctly specified, its estimates can be highly inaccurate when either of these conditions is not met \citep{huang2023learning, schmitt2023leveraging, wehenkel2024addressing, mishra2025robust}. Consequently, training solely on samples from the prior predictive distribution may be insufficient to ensure robust posterior estimates. Since Bayesian self-consistency losses described below can be applied to empirical data rather than simulated data from the prior predictive distribution, they are suitable to overcome these limitations of standard simulation-based training.

\subsection{Bayesian Self-Consistency Losses} \label{sec:sc}

Bayesian self-consistency (SC) losses \citep{schmitt2023leveraging, mishra2025robust, kucharsky2025towards, kothe2023review, richter2020vargrad, richter2023improved, sanokowski2025rethinking} are strictly-proper scores that leverage a simple symmetry in Bayes’ rule to enforce accurate posterior estimation: Under exact inference, the marginal likelihood is independent of the parameters $\theta$. In other words, the Bayesian self-consistency ratio between the likelihood-prior product and the posterior remains constant across any set of parameter values $\theta^{(1)}, \dots, \theta^{(L)}$,
\begin{equation}
    p(x) = \frac{p(x \mid \theta^{(1)})p(\theta^{(1)})}{p(\theta^{(1)} \mid x)}
    = \dots = \frac{p(x \mid \theta^{(L)})p(\theta^{(L)})}{p(\theta^{(L)} \mid x)}.
\end{equation}
However, when the true posterior $p(\theta \mid x)$ is replaced with a neural approximation $q_{\Phi}(\theta \mid x)$, bespoke ratios will longer be constant for any imperfect approximation. 
We can equivalently say that the log ratios have positive variance, which readily implies a loss function to be minimized. We will use the unsupervised formulation of the (variance-based) self-consistency loss, which utilizes unlabeled real-world data \citep{mishra2025robust}. Hereafter, we refer to the variance-based self-consistency loss simply as the ``self-consistency loss'' or short ``SC loss''.
For $M$ instances of real-world data $x^*_m \sim p^*(x)$, the SC loss is given by
\begin{equation}
\label{eq:sc-loss}
\begin{aligned}
\mathcal{L}_{\text{SC}}(\Phi) 
&= \mathcal{L}_{\text{SC}}(\Phi; x^*) \\
&= \frac{1}{M} \sum_{m=1}^M 
   \operatorname{Var}_{\theta \sim p_C(\theta)} \Big[ 
   \log p(x_m^* \mid \theta) 
       + \log p(\theta) \\
&\quad\quad - \log q_{\phi}(\theta \mid h_{\psi}(x_m^*)) 
   \Big].
\end{aligned}
\end{equation}
Here, $p_C(\theta)$ is a parameter proposal distribution, which we set to the current approximate posterior $q_{\phi, r}(\theta \mid h_{\psi, r}(x_m^*))$ at training iteration $r$ \citep{mishra2025robust}. 

The SC loss possesses several properties that make it particularly appealing for use in CL scenarios. First, it does not require labeled data, allowing the use of real-world unlabeled data to minimize the SC loss. Second, the SC loss provides additional signal which leads to more accurate posterior estimation even for atypical data \citep{mishra2025robust, kucharsky2025towards}, where pure simulation-based training fails \citep{frazier2024statistical, ward2022robust, mishra2025robust, schmitt2023detecting}. 
Third, the SC loss can attain accurate posterior approximations with comparatively little training data, making it data-efficient. 

\subsection{Regimes of Self-Consistency training}
\label{sec:sc_all}

Prior work has used SC losses in combination with SB losses in a semi-supervised setup during training \citep{mishra2025robust, kucharsky2025towards, schmitt2023leveraging}. In a CL setting, tasks arrive sequentially $\{\mathcal{T}_1, \dots, \mathcal{T}_T\}$ and observed data from different tasks may not be available at the same time. Further, because the approximate posterior is initially highly inaccurate, the SC loss tends to be numerically unstable at first; this often requires careful weight scheduling of the two losses for stable training \citep{mishra2025robust, schmitt2023leveraging, kucharsky2025towards}.

To address these issues, we split training into two stages. First, we perform standard simulation-based pre-training of the posterior network using the SB loss. After this initial step, the SB loss is no longer used. Second, we train the posterior network using the SC loss as we accrue data from different tasks. Next, we describe different training regimes for using the SC loss on new data, regimes that constitute the main focus of this article.

\subsubsection{Test-time Training with Self-Consistency} 
\label{sec:test-time}

In the test-time training regime, as data from new tasks arrive, the simulation-based pre-trained network is further trained with the SC loss on data from the current task only. This is the closest method to the combined SB and SC loss studied in recent work \citep{mishra2025robust, kucharsky2025towards}. No CL mechanism is employed; updates are performed solely to improve posterior estimates for the current task, without explicitly retaining performance on other tasks. This setting is particularly advantageous in scenarios where accuracy on the current task is paramount and long-term retention across tasks is not required.

\subsubsection{Continual Learning with Self-Consistency} 
\label{sec:naive-sc}

In a CL regime, we use the SC training in a sequential fashion, where at step $t$, we minimize the SC loss using data from task $\mathcal{T}_t$ for a network that was pre-trained by the SC loss for all previous tasks $\{\mathcal{T}_1,\dots, \mathcal{T}_{t-1}\}$ sequentially (and initially with the SB loss). Crucially, because the SC objective does not explicitly constrain the network on earlier task distributions, updates driven solely by the current task can overwrite representations important for previous tasks. As we later show, this may lead to catastrophic forgetting, and we further call this approach ``naive SC''.

\subsubsection{Self-Consistency with Episodic Replay} \label{sec:sc-er}

To alleviate catastrophic forgetting in CL, we employ an episodic replay (ER) mechanism \citep{rolnick2019experience, chaudhry2019continual, shin2017continual} that stores unlabeled real-world data from previous tasks $\{\mathcal{T}_1, \mathcal{T}_2, \ldots, \mathcal{T}_{t-1}\}$ in a memory buffer $\mathcal{E}$. During training with self-consistency loss on a new task $\mathcal{T}_t$, the network is jointly optimized on the current task data and stored samples from $\mathcal{E}$, encouraging a stability--plasticity balance. This allows the network to retain knowledge of past tasks while adapting to new tasks. The resulting episodic replay based self-consistency (SC-ER) loss is given by
\begin{equation}
    \mathcal{L}_{\text{SC-ER}}(\Phi) = \mathcal{L}_{\text{SC}}(\Phi; x^{*}_{SC}), 
    \label{eq:sc-er}
\end{equation}
where $x^{*}_{SC} = x_t^{*} \cup \mathcal{E}$, $x_t^{*}$ represents the data from the  task $\mathcal{T}_t$, and $\mathcal{E}$ contains a subset of data from previous tasks. An additional benefit of using ER is that it preserves the inference target of self-consistency (analytic posterior) \citep{elsemuller2025does}, because including replay buffer within the same loss function does not involve any regularization term unlike other CL approaches \citep{kirkpatrick2017overcoming, zenke2017continual, aljundi2018memory, li2017learning}.

As the number of tasks increase, storing even a few representative samples from each task may become impractical \citep{rolnick2019experience, li2024adaer}. To address this, we maintain a minimal replay buffer that initially stores few observations per task. Once the memory capacity is reached, we apply K-medoids clustering \citep{kaufman2009finding} over all stored samples and retain only the cluster representatives. 
The number of clusters is set to match the maximum buffer size, ensuring that the replay memory remains within the predefined limit while preserving diversity among stored observations. This diversity preserving approach is suitable to be used in combination with SC as it works on minimal data available per task. The overall procedure is summarized in Algorithm \ref{alg:sc-episodic-replay}.

\begin{algorithm}[h]
\caption{Self-Consistency with Episodic Replay}
\label{alg:sc-episodic-replay}
\begin{algorithmic}[1] 

\STATE \textbf{Input:} Simulated dataset $(\theta, x) \sim p(\theta, x)$, 
unlabeled real-world tasks $\{\mathcal{T}_t\}_{t=1}^T$, replay buffer size $K$
\STATE \textbf{Output:} Trained posterior network $q_\Phi(\theta \mid x)$

\STATE Initialize $q_\Phi(\theta \mid x)$, replay buffer $\mathcal{E} \leftarrow \emptyset$

\STATE \textbf{Simulation based pre-training}
\FOR{each pretraining iteration}
    \STATE Sample minibatch $(\theta, x) \sim p(\theta, x)$
    \STATE Update $q_\Phi$ by minimizing Eq.~\ref{eq:npe}
\ENDFOR

\STATE \textbf{CL on real-world tasks}
\FOR{$t = 1$ to $T$}
    \STATE Obtain current dataset $x_t^* \sim p_t^*(x)$ from task $\mathcal{T}_t$

    \STATE Concatenate replay data with current data: \\
        $x^{*}_{\text{SC}} \leftarrow x_t^{*} \cup \mathcal{E}$

    \FOR{each SC-ER iteration}
        \STATE Compute SC-ER loss $\mathcal{L}_{\text{SC-ER}}$ as in Eq. \ref{eq:sc-er}
        \STATE Update $q_{\Phi}$ by minimizing $\mathcal{L}_{\text{SC-ER}}$
    \ENDFOR    
    \IF{using clustering}
       \STATE Draw posterior samples $\{\theta^{(s)}_t\}$ from $q_\Phi(\theta \mid x_t^*)$
       \STATE $\mathcal{E} \leftarrow$ \textsc{KMedoidUpdate}$(\mathcal{E}, x_t^*, \{\theta^{(s)}_t\})$: Alg. \ref{alg:k-medoid}
    \ELSE
       \STATE Store a subset of data from $\mathcal{T}_t$ in $\mathcal{E}$
    \ENDIF
\ENDFOR

\STATE \textbf{Return} $q_\Phi(\theta \mid x)$

\end{algorithmic}
\end{algorithm}

\subsubsection{Self-Consistency with Elastic Weight Consolidation} 

In certain CL settings, retaining data from previous tasks is impractical due to privacy constraints \citep{tong2025model} or long task sequences that make replay inefficient as the memory buffer grows linearly with number of tasks \citep{shin2017continual, rolnick2019experience}. 
%
%
In such cases, we propose to use weight regularization, a widely used strategy to mitigate catastrophic forgetting in continual  learning \citep{kirkpatrick2017overcoming, zenke2017continual, aljundi2018memory}. The core idea is to constrain updates to network parameters that are important for previously learned tasks while allowing flexibility for adapting to new tasks. This is achieved by introducing a regularization term in the loss function that penalizes large deviations of the current network parameters from their previously learned values.
Formally, after training on task $\mathcal{T}_{i}$, the optimized network parameters are denoted by $\Phi_{i}^*$. When learning a new task $\mathcal{T}_t$, we minimize the following objective:
\begin{equation}
\label{eq:weight-reg}
\mathcal{L}_{\text{SC-WR}}(\Phi_{t}) = \mathcal{L}_{\text{SC}}(\Phi_{t}) + \lambda \ \sum^{t-1}_{i=1} \ \Omega(\Phi_{t}, \Phi_{i}^*),
\end{equation}
where $\mathcal{L}_{\text{SC}}$ denotes the task-specific SC loss for $\mathcal{T}_t$, 
$\Omega(\cdot)$ is a regularization term that measures the deviation of the current parameters $\Phi_{t}$ from the previous parameters $\Phi_{i}^*$, and $\lambda$ is a hyperparameter controlling the trade-off between retaining previous knowledge and adapting to new tasks. 
We instantiate $\Omega(\cdot)$ as quadratic penalty taking inspiration from Elastic Weight Consolidation~\citep[EWC,][]{kirkpatrick2017overcoming}:
\begin{equation}
\label{eq:ewc}
\Omega(\Phi_{t}, \Phi_{i}^*) =  \sum_{j} \mathcal{W}_{jj} \, (\Phi_{t,j} - \Phi_{i,j}^*)^2,
\end{equation}
where $\mathcal{W}_{jj}$ represents the diagonal importance matrix that estimates the importance of parameter $\Phi_{i,j}^*$ for previous task $\mathcal{T}_i$, and $j$ indexes all trainable network parameters. 
%
%
After convergence on task $\mathcal{T}_i$, we compute the diagonal information matrix as, 
\begin{equation}
\label{eq:fisher-sc}
\mathcal{W}_{jj}^{(i)} =
\mathbb{E}_{x^* \sim p_i^*(x)} 
\left[
\left(
\frac{\partial \mathcal{L}_{SC}(x^*)}{\partial \Phi_j}
\right)^2
\right],
\end{equation}
\begin{equation}
\label{eq:fisher-sc-exp}
\mathcal{W}_{jj}^{(i)} \approx 
\frac{1}{K} \sum_{k=1}^{K}
\left[
\left(
\frac{\partial \mathcal{L}_{SC}(x_{k}^{*})}{\partial \Phi_j}
\right)^2
\right].
\end{equation}
where the expectation in Eq. \ref{eq:fisher-sc} is approximated from $K$ unlabeled samples from $\mathcal{T}_i$ in Eq. \ref{eq:fisher-sc-exp}. This approach effectively anchors critical parameters while allowing less important ones to adapt freely. As we utilize the SC loss, it allows us to compute parameter importance using only unlabeled real-world data. 
When training the network on task $\mathcal{T}_t$, we minimize the following SC-EWC loss: 
\begin{equation}
    \mathcal{L}_{\text{SC-EWC}}(\Phi_{t}) = \mathcal{L}_{\text{SC}}(\Phi_{t};x_{t}^*) + \frac{\lambda}{2}  \ \sum^{t-1}_{i=1} \sum_{j} \ \mathcal{W}_{jj}^{(i)}(\Phi_{t,j} - \Phi_{i,j}^*)^2
    \label{eq:sc-ewc}
\end{equation}

The overall procedure is summarized in Algorithm \ref{alg:sc-ewc}. We also combine ER and EWC with self-consistency simultaneously to see their combined effects (see Algorithm \ref{alg:sc-er-ewc}). In such a case, the loss becomes:
\begin{equation}
\begin{aligned}
\mathcal{L}_{\text{SC-ER-EWC}}(\Phi_{t}) 
&= \mathcal{L}_{\text{SC-ER}}(\Phi_{t};x_{SC}^*) \\
&\quad + \frac{\lambda}{2} \sum_{i=1}^{t-1} \sum_{j} 
\mathcal{W}_{jj}^{(i)}(\Phi_{t,j} - \Phi_{i,j}^*)^2
\end{aligned}
\label{eq:sc-er-ewc}
\end{equation}

\begin{algorithm}[h]
\caption{SC with Elastic Weight Consolidation}
\label{alg:sc-ewc}
\begin{algorithmic}[1]

\STATE \textbf{Input:} Simulated dataset $(\theta, x) \sim p(\theta, x)$, unlabeled tasks $\{\mathcal{T}_t\}_{t=1}^T$, EWC weight $\lambda$, Fisher batches $B$
\STATE \textbf{Output:} Trained posterior network $q_{\Phi}(\theta \mid x)$

\STATE Initialize posterior $q_{\Phi}$, EWC record list $\mathcal{R} \leftarrow \emptyset$

\STATE \textbf{Simulation-based pre-training}
\FOR{pretraining iterations}
    \STATE Sample $(x, \theta) \sim p(\theta, x)$
    \STATE Update $q_\Phi$ by minimizing Eq.~\ref{eq:npe}
\ENDFOR

\STATE \textbf{CL}
\FOR{$t = 1$ to $T$}
    \STATE Obtain current dataset $x_t^* \sim p_t^*(x)$ from task $\mathcal{T}_t$
    
    \FOR{each SC-EWC iteration}
        
        \IF{$\mathcal{R} \neq \emptyset$}
            \STATE Update $q_\Phi$ by minimizing $\mathcal{L}_{\text{SC-EWC}}$ from Eq. \ref{eq:sc-ewc}
        \ELSE
            \STATE Update $q_\Phi$ by minimizing $\mathcal{L}_{\text{SC}}$ from Eq. \ref{eq:sc-loss}
        \ENDIF
    \ENDFOR
    
    \STATE Estimate $\mathcal{W}_{jj}^{{(t)}}$ for task $\mathcal{T}_t$ as per Eq.~\ref{eq:fisher-sc-exp}
    \STATE Snapshot parameters $\Phi_{t}^*$
    \STATE Append $(\Phi_{t}^*, \mathcal{W}_{jj}^{(t)})$ to EWC record list $\mathcal{R}$
\ENDFOR

\STATE \textbf{Return} $q_{\phi}(\theta \mid x)$

\end{algorithmic}
\end{algorithm}

\section{Related Work} \label{sec:related}

CL has been extensively studied in the last decades as overcoming catastrophic forgetting remains a major challenge in deep learning. Three major families of approaches have emerged: (1) Regularization-based methods \citep{kirkpatrick2017overcoming, zenke2017continual, aljundi2019task, kozal2024continual, zhao2024statistical, lewandowski2025learning, gomez2024plasticity} constrain important network parameters from changing too much. (2) Replay-based methods \cite{lee2019robust, rolnick2019experience, chaudhry2019continual, aghasanli2025prototype, yoo2024layerwise, madaan2021representational} combine current data with stored or generated samples from past  during training. (3) Parameter-isolation and architectural approaches \citep{rusu2016progressive, lu2024revisiting, schwarz2018progress, rao2019continual} allocate dedicated sub-networks or dynamically expand networks to prevent forgetting. In this work, we utilize a replay and a regularization based CL approach. We do not include architecture expansion methods, as their parameter growth scales with the number of tasks \citep{rusu2016progressive}. Also, such approaches typically restrict parameter sharing across tasks, limiting positive transfer when tasks are related. 


Recent work has increasingly investigated robustness in ABI and, more broadly, simulation-based inference \citep{dellaporta2022robust, frazier2024statistical, gloeckler2023adversarial}. These approaches can be broadly grouped into two categories: (i) methods that analyze or detect discrepancies between simulated and real-world data \citep{schmitt2023detecting, alvey2026simulation, cannon2022investigating}, and (ii) methods that directly mitigate the impact of simulation gap on posterior estimation \citep{ward2022robust, huang2023learning, gloeckler2023adversarial, wehenkel2024addressing, mishra2025robust, elsemuller2025does, kucharsky2025towards}. Our work falls into the latter category. Within this line of research, regularization-based approaches \citep{gloeckler2023adversarial, huang2023learning} introduce additional constraints to correct the posterior estimates. Other approaches such as \cite{wehenkel2024addressing} incorporate limited real-world labeled data to supplement the simulation-based training, or apply unsupervised domain adaptation techniques to handle distribution shifts \citep{elsemuller2025does}. More recently, SC-based approaches enhance robustness under model misspecification with unlabeled real-world data \citep{mishra2025robust, kucharsky2025towards}. 

To the best of our knowledge, no prior work has explored CL for ABI, a gap that we address. Existing methods in ABI assume either simultaneous availability of real-world data at training time or that future tasks come from the same distribution, thereby lacking the mechanisms to prevent catastrophic forgetting. By explicitly formulating unsupervised CL within the ABI framework and leveraging SC alongside replay and regularization-based CL strategies, our approach effectively mitigates catastrophic forgetting and improves posterior estimation across sequential tasks.

\section{Experiments} \label{sec:exp}

\paragraph{Experimental setup} We conduct three experiments ranging from linear regression to time-series and racing diffusion models. For each experiment, we initially perform simulation-based training of the posterior network using the SB loss in an online setting, which also acts as a baseline. We then train the posterior network sequentially across all tasks in an unsupervised CL setting using our proposed methods (SC-ER, SC-EWC, and SC-ER-EWC) along with naive SC. After completion of training on the final task, we evaluate every method on all the tasks. To assess the capacity of the network for accurate performance on each individual task, we contrast results of the CL methods with test-time SC trained on each task. This directly reveals the extent to which each method mitigates (or fails to mitigate) catastrophic forgetting. All experiments are conducted with the Python library BayesFlow \citep{kuhmichel2026bayesflow}.

\paragraph{Metrics} We use MCMC-based posterior estimates via Stan \citep{carpenter2017stan} as a non-amortized gold-standard throughout our experiments. We measure mean bias as the difference between the posterior mean estimated by different methods and the posterior mean obtained using Stan. Analogously, standard deviation bias quantifies discrepancies in posterior uncertainty. We also compute the Maximum Mean Discrepancy \citep[MMD;][]{gretton2012kernel} between posterior samples from our method and those from Stan, using a Gaussian kernel. For better interpretability, we report the ratio of MMD of a specific method to MMD of the SB training baseline. MMD ratio values smaller than one indicate improved performance compared to the baseline.

\subsection{Experiment 1: Linear Regression} \label{sec:lin_reg}

\begin{figure*}[h]
    \centering
    \includegraphics[width=0.99\linewidth]{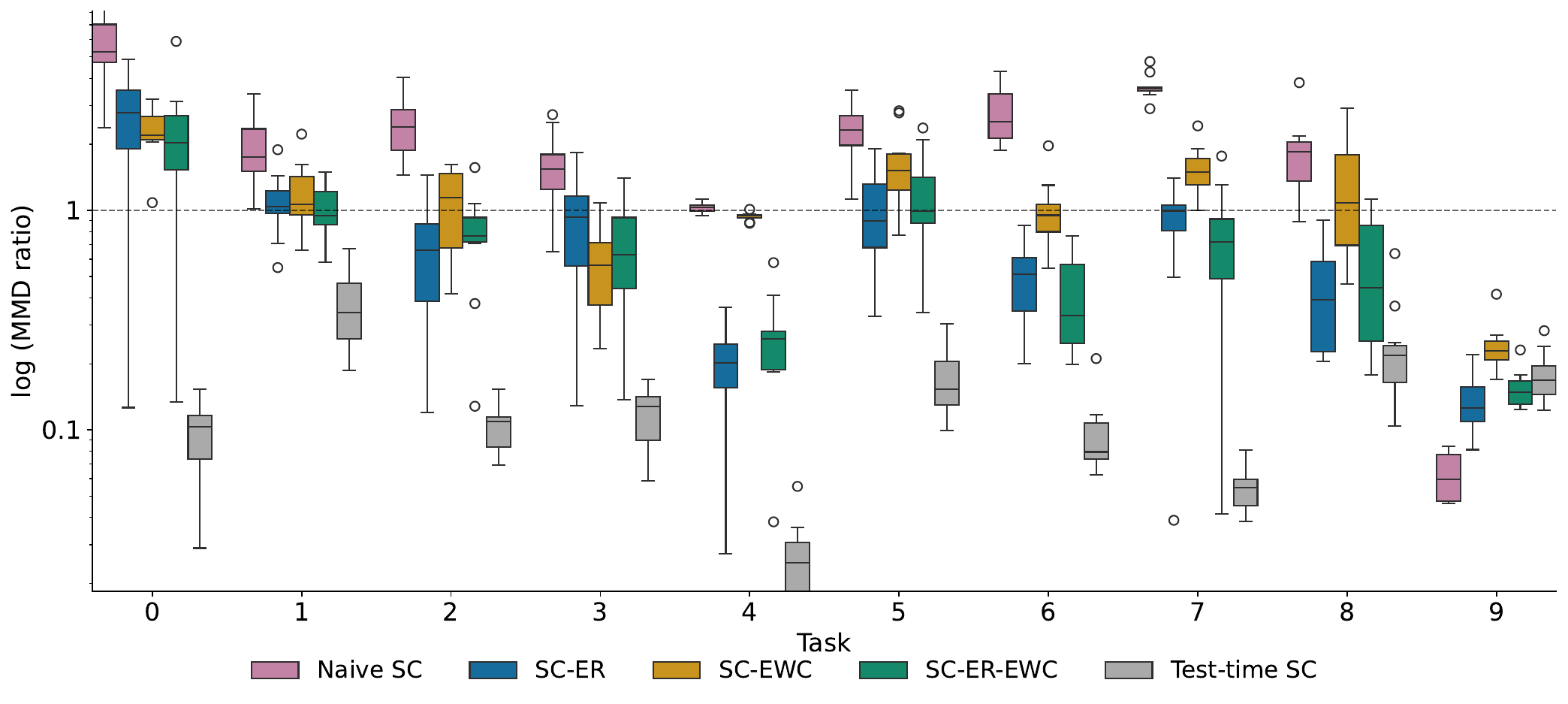}
    \caption{\textbf{Experiment 1.} MMD ratio in log scale across CL tasks (0–9) for different methods. Boxes summarize variability  across subsets. Dashed line marks parity with simulation-based (SB) baseline (ratio = 1). Naive SC shows catastrophic forgetting in CL setting whereas our proposed methods mitigate forgetting and provide better posterior estimates compared to SB and Naive SC. Test-time SC also gives accurate posterior estimates.}
    \label{fig:lin_reg}
\end{figure*}

First, we consider a standard Bayesian linear regression model with five predictors and $N=100$ observations. The generative model is
\begin{equation}
    y_n \sim \mathcal{N}\left(\alpha + \mathbf{x}_n^\top \boldsymbol{\beta}, \sigma^2\right),
\end{equation}
where $\alpha$ is the intercept, $\boldsymbol{\beta} \in \mathbb{R}^5$ are the regression coefficients, and $\sigma$ is the observation noise. We place independent priors on all parameters:
\begin{equation}
\begin{aligned}
\alpha &\sim \mathcal{N}(0,1), \qquad \sigma \sim \mathrm{HalfNormal}(1),\\[6pt]
\boldsymbol{\beta} &\sim \mathcal{N}\!\left(\mathbf{0},\,\mathrm{diag}(\tau^2)\right), \qquad
\tau = (1,1,1,1,1).
\end{aligned}
\end{equation}
The above model produces a diagonal covariance structure for the slope parameters. Training was performed as described in the experimental setup. For CL, we utilize nine real-world publicly available datasets (see Appendix \ref{sec:lr_dataset}) which constitute 9 CL tasks $\{\mathcal{T}_1, ..., \mathcal{T}_9\}$. We further divided these datasets in ten subsets each having $N = 100$ observations. We also utilize one simulated dataset $\{\mathcal{T}_0\}$, where the ten subsets are independently drawn from prior predictive distribution. For SC-ER replay buffer, we only utilize one subset per task.  

\paragraph{Results} In Figure~\ref{fig:lin_reg}, we report the MMD ratio across the ten tasks for naive SC, test-time SC, and the proposed SC-ER, SC-EWC, and SC-ER-EWC methods. These results show that naive SC suffers from severe catastrophic forgetting across tasks and even performs worse than the SB baseline in the CL setting. In contrast, all proposed CL approaches mitigate catastrophic forgetting and yield posterior estimates that are significantly closer to the reference posterior than both naive SC and the SB baseline for most tasks. 
Among the proposed methods, SC-ER provides the most accurate posterior estimates across all tasks. It effectively prevents catastrophic forgetting while requiring only a single subset per task as a replay buffer, demonstrating that the SC loss can produce accurate posterior estimates with minimal retained data. SC-EWC outperforms naive SC and mitigates forgetting, although its performance remains inferior to SC-ER and only marginally better than SB baseline. 

The combined SC-ER-EWC approach likewise mitigates catastrophic forgetting. Its performance is comparable to SC-ER and better than SC-EWC. This suggests that incorporating episodic replay within the self-consistency objective is already sufficient to preserve performance across tasks, and adding the EWC penalty does not provide much additional gains in this setting. Overall, these results highlight the effectiveness of incorporating self-consistency with lightweight replay mechanisms for robust posterior estimation in CL scenarios. Test-time SC gives accurate posterior estimates when current task inference is the only objective rather than overcoming catastrophic forgetting. Test-time SC prioritizes plasticity (rapid adaptation to the current task) at the expense of stability. In contrast, our CL methods balance plasticity and stability, maintaining performance across all tasks. Additional results including hyperparameter sensitivity analysis for SC-EWC and SC-ER-EWC are provided in Appendix~\ref{sec:app_res1}.

\subsection{Experiment 2: Air Passenger Traffic Forecasting} \label{sec:air}

\begin{figure}[h]
    \centering
    \includegraphics[width=\linewidth]{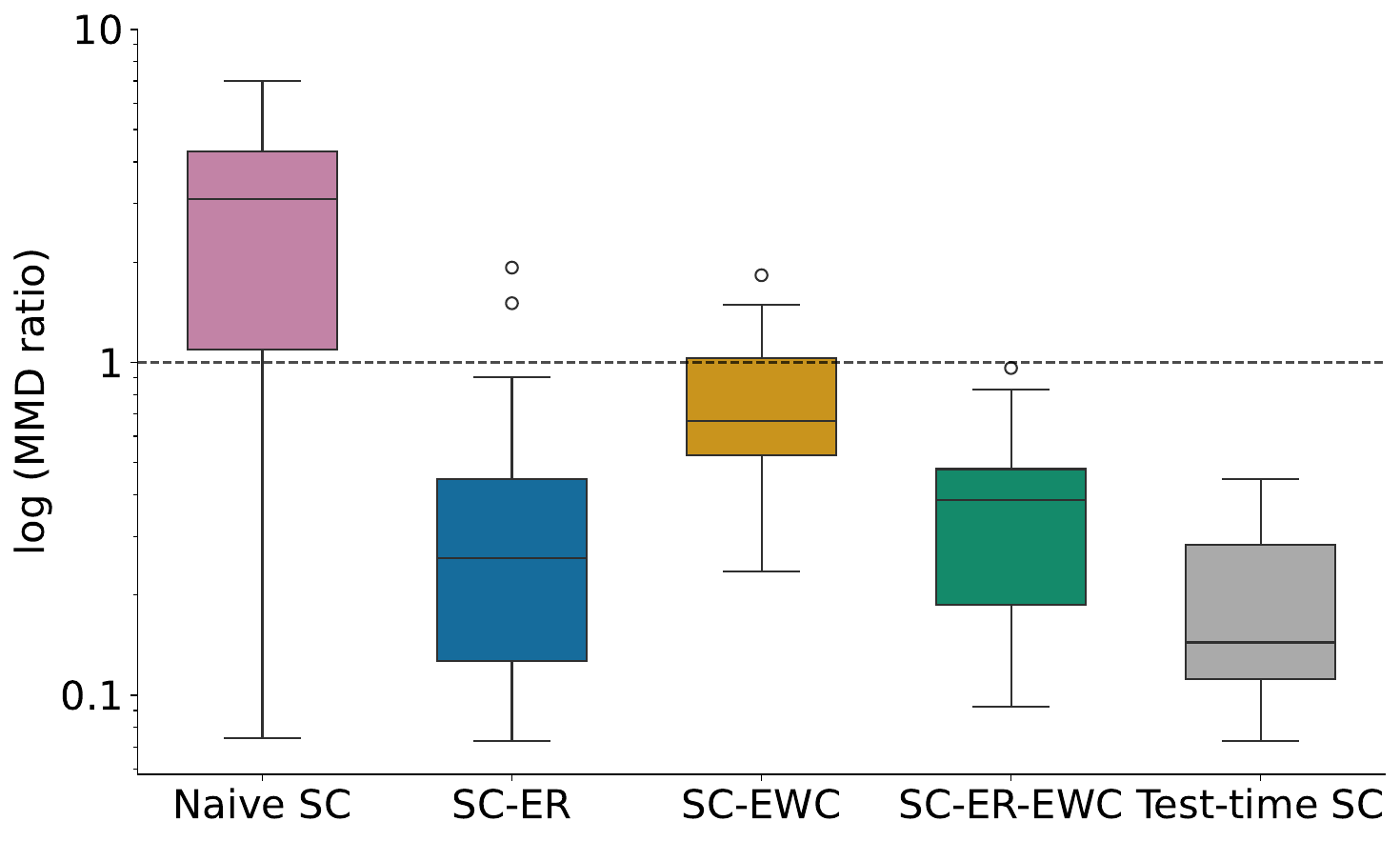}
    \caption{\textbf{Experiment 2.} MMD ratio in log scale for different methods aggregated over fifteen CL tasks.  Dashed line marks parity with simulation-based baseline (ratio = 1). Naive SC shows catastrophic forgetting in CL setting whereas our proposed methods mitigate forgetting and perform better than both SB baseline and naive SC.
    }
    \label{fig:air_mmd}
\end{figure}

Second, we analyze trends in European air passenger traffic data provided by \citet{Eurostat2022a, Eurostat2022b, Eurostat2022c}. This is a widely used case study in the ABI literature \citep{habermann2024amortized, mishra2025robust, kucharsky2025towards}. We retrieve time series of annual air passenger counts between 15 European countries (departures) and the USA (destination) from 2004 to 2019. These 15 countries constitute 15 CL tasks. We fit the following auto-regressive process of order 1:
\begin{equation}
  y_{j, t+1} \sim \mathcal{N}(\alpha_j + y_{j, t}\beta_j + u_{j, t}\gamma_j + w_{j,t}\delta_j, \sigma_j),
\end{equation}
where the target quantity $y_{j, t+1}$ is the difference in air passenger traffic for country $j$ between time $t + 1$ and $t$. To predict $y_{j, t+1}$ we use two additional predictors: $u_{j, t}$ is the annual household debt of country $j$ at time $t$, measured in \% of gross domestic product (GDP) and $w_{j,t}$ is the real GDP per capita. Additional details about the dataset and model are given in Appendix~\ref{sec:app_air_model}. We train the neural posterior estimators for all the methods as described in the experimental setup. For episodic replay buffer, we only utilize the data of eight tasks which are chosen based on K-medoid clustering described in Section~\ref{sec:sc-er}.

\paragraph{Results} In Figure~\ref{fig:air_mmd}, we present the MMD ratio, aggregated over the 15 tasks for naive SC, test-time SC, and the proposed SC-ER, SC-EWC, and SC-ER-EWC methods. Figure~\ref{fig:air_mmd_full} in Appendix~\ref{sec:air_results} depicts the MMD ratio across tasks. Similar to experiment~\ref{sec:lin_reg}, naive SC exhibits substantial catastrophic forgetting, whereas the proposed CL approaches effectively mitigate forgetting across tasks. SC-ER provides the most accurate posterior estimates throughout all tasks while maintaining a memory buffer containing data from eight countries for episodic replay. We further investigate the effect of varying the number of tasks retained in the replay buffer which is reported in Figure~\ref{fig:k-med}, Appendix~\ref{sec:air_results}. SC-EWC also overcomes catastrophic forgetting and gives better posterior estimates compared to both naive SC and SB baseline. However, it performs slightly worse that SC-ER. 

We observe that SC-ER-EWC performs better than SC-EWC and comparable to SC-ER, indicating that combining ER and EWC can further stabilize learning, though replay appears to be the dominant contributing factor. We conducted additional experiments varying the size of the replay buffer and observe that the performance of SC-ER-EWC degrades at low memory budget. It nonetheless consistently outperforms both baseline SB and naive SC, as shown in Figure~\ref{fig:k-med-er-ewc}, Appendix~\ref{sec:air_results}. Test-time SC gives accurate posterior estimates when current task inference is the only objective rather than overcoming catastrophic forgetting. However, for most of the tasks, SC-ER performs on par with test-time SC. Additional results are described in Appendix~\ref{sec:air_results}.

\subsection{Experiment 3: Racing Diffusion of Decision Making} 
\label{sec:diffusion}

\begin{figure*}[h]
    \centering
    \includegraphics[width=0.99\linewidth]{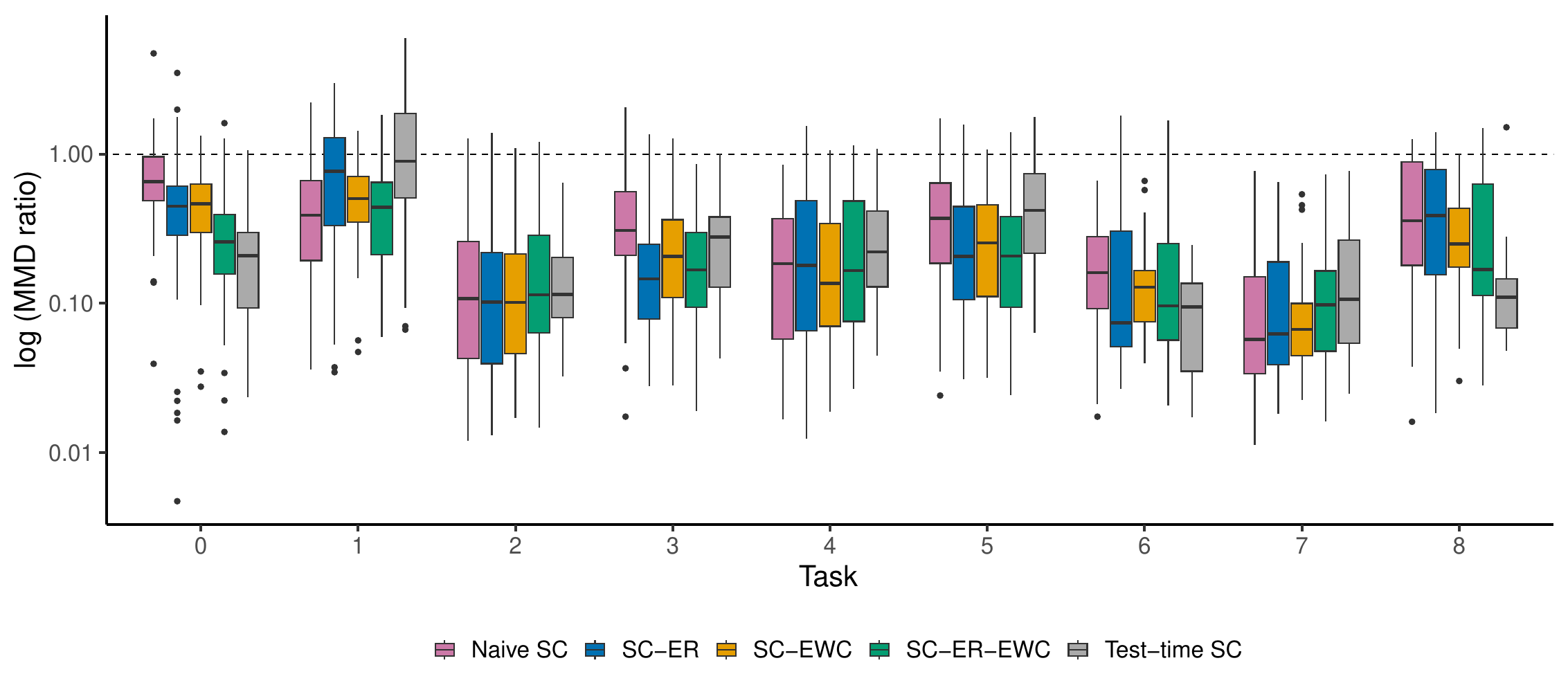}
    \caption{\textbf{Experiment 3.} MMD ratio in log scale across CL tasks (0–8) for the four RDM parameters for different methods. Boxes summarize variability over datasets. Dashed line marks parity with simulation-based baseline (ratio = 1). All SC methods outperform SB-only training. Catastrophic forgetting is not evident.}
    \label{fig:diffusion_mmd_ratio}
\end{figure*}

Lastly, we analyze data sets available from a large corpus of experiments on attentional control \citep{haaf2025attentional}. From this database, we extracted eight data sets. These eight data sets constitute the eight CL tasks. We further generated one simulated dataset $\{\mathcal{T}_0\}$ of 50 observations from the prior predictive distribution. The data consists of response times (in msec) and response (correct/incorrect).

In this case study, we use the racing diffusion model \citep[RDM,][]{tillman2020sequential}. The model assumes that each response is associated with a dedicated noisy evidence accumulator. Each accumulator begins at zero and evolves according to a Wiener process with drift (where $W_t$ is the standard Wiener process),
\begin{equation}
X_t = \nu~\mathrm{d}t + \sigma~\mathrm{d}W_t,
\qquad X_0 = 0.
\end{equation}
A response is initiated when one of the accumulators first reaches the decision boundary $\alpha$. The observed response time is the hitting time of the winning accumulator plus a non-decision time $\tau$. The parameters of the model are the decision boundary $\alpha$, non-decision time $\tau$, and two drift rates $\nu_1$ and $\nu_2$ (associated with correct and incorrect responses, respectively).

We train the neural posterior estimators for all methods as described in the experimental setup. For episodic replay buffer, we retain data from five participant from each previous task -- no clustering has been done in this case study. More details about the data, model, and training are in Appendix~\ref{sec:app_diffusion}.

\paragraph{Results} Figure~\ref{fig:diffusion_mmd_ratio} shows the MMD ratio across all 9 tasks. In contrast to previous experiments, naive SC does not exhibit pronounced catastrophic forgetting; all methods that utilize any SC training perform clearly better than the SB baseline. Further, CL methods perform on-par even with test-time SC. One possible interpretation of these findings is that the severity of forgetting is dependent on the variability of data within tasks relative to the variability of data between tasks. In the current case study, even though the data sets come from different experiments, they all exhibit similar characteristics: average accuracy ranging between 90\%--95\% with mean response times between 0.4--0.7 sec. Therefore, it may be the case that the datasets between tasks are similar enough that catastrophic forgetting is not an issue in the first place.

\section{Discussion}

ABI enables fast posterior inference after simulation-based training, but its accuracy can degrade under distribution shifts. Bayesian self-consistency (SC) losses improve robustness of ABI by leveraging unlabeled empirical data, yet prior work assumes static or single-task setting where all empirical data is readily available. In this work, we adopt a two-stage training procedure consisting of simulation-based pre-training followed by unsupervised self-consistency fine-tuning, to enable network adaptation as new data come in. 

We demonstrate that naive application of this procedure can lead to catastrophic forgetting, as network adaptation to a new task can overwrite representations important for earlier tasks, degrading accuracy of the posterior estimates. To address this, we introduce more elaborate continual learning mechanisms. Episodic replay (SC-ER) stabilizes learning by retaining a small set of past observations, while elastic weight consolidation (SC-EWC) provides an alternative that constrains parameter updates using importance estimates derived from SC gradients. Across three diverse case studies, both approaches substantially reduce forgetting and yield posterior estimates closer to MCMC reference than simulation-only training and naive SC. The experiments further indicate that the severity of forgetting depends on the degree of distributional variation between tasks. This points to a practical takeaway: when tasks are heterogeneous, replay and/or regularization become important to maintain performance across the full task history.

\paragraph{Limitations and future direction} Our self-consistency based continual learning approaches rely on fast density evaluations during training. This makes free-form methods such as flow matching \citep{lipman2022flow} or score-based diffusion \citep{song2020denoising} less practical due to their need for numerical integration.

The employed self-consistency losses assume access to the analytic prior and likelihood densities of the probabilistic model. With implicit prior or likelihoods, self-consistency loss is not strictly proper \citep{mishra2025robust} and training with self-consistency has smaller benefits for the accuracy of the posterior approximation \citep{mishra2025robust,kucharsky2025towards}. Whether self-consistency or other unsupervised losses can be used effectively in a continual learning setting without access to explicit prior and likelihood densities remains a topic for future research.

\begin{contributions} 
    Aayush Mishra and Paul Bürkner conceived the idea. Aayush Mishra developed the methods with the help of Paul Bürkner. Aayush Mishra implemented the first two experiments. \v{S}imon Kucharsk\'{y} implemented the third experiment. Aayush Mishra and \v{S}imon Kucharsk\'{y} drafted the article. 
    All authors contributed to final editing of the article.
\end{contributions}

\begin{acknowledgements} 
    Paul Bürkner acknowledges support of the DFG Collaborative Research Center 391 (Spatio-Temporal Statistics for the Transition of Energy and Transport) -- 520388526.
\end{acknowledgements}

\bibliography{references}

\newpage

\onecolumn

\title{Unsupervised CL for Amortized Bayesian Inference\\(Supplementary Material)}
\maketitle

\appendix
\section{Additional algorithms} \label{sec:app_alg}

\begin{algorithm}[h]
\caption{Self-Consistency with Episodic Replay and Elastic Weight Consolidation}
\label{alg:sc-er-ewc}
\begin{algorithmic}[1]

\STATE \textbf{Input:} Simulated dataset $(\theta, x) \sim p(\theta, x)$, unlabeled tasks $\{\mathcal{T}_t\}_{t=1}^T$, EWC weight $\lambda$, Fisher batches $B$, replay size $K$
\STATE \textbf{Output:} Posterior estimator $q_{\Phi}(\theta \mid x)$

\STATE Initialize $q_{\Phi}$, replay buffer $\mathcal{E} \leftarrow \emptyset$, EWC record list $\mathcal{R} \leftarrow \emptyset$

\STATE \textbf{Simulation-based pre-training}
\FOR{pretraining iterations}
    \STATE Sample $(\theta, x) \sim p(\theta, x)$
    \STATE Update $\Phi$ by minimizing $\mathcal{L}_{SB}$ from Eq. \ref{eq:npe}
\ENDFOR

\STATE \textbf{CL}
\FOR{$t = 1$ to $T$}
    \STATE Obtain current dataset $x_t^* \sim p_t^*(x)$ from task $\mathcal{T}_t$
    \STATE Concatenate replay data with current data: \\
        $x^{*}_{\text{SC}} \leftarrow x_t^{*} \cup \mathcal{E}$
    
    \FOR{each SC-ER-EWC iteration}
        
        \IF{$\mathcal{R} \neq \emptyset$ and $\mathcal{E} \neq \emptyset$}
            \STATE Update $q_\Phi$ by minimizing $\mathcal{L}_{\text{SC-ER-EWC}}$ from Eq. \ref{eq:sc-er-ewc}
        \ELSE
            \STATE Update $q_\Phi$ by minimizing $\mathcal{L}_{\text{SC}}$ from Eq. \ref{eq:sc-loss}
        \ENDIF

    \ENDFOR
    \IF{using clustering}
       \STATE Draw posterior samples $\{\theta^{(s)}_t\}$ from $q_\Phi(\theta \mid x_t^*)$
       \STATE $\mathcal{E} \leftarrow$ \textsc{KMedoidUpdate}$(\mathcal{E}, x_t^*, \{\theta^{(s)}_t\})$
    \ELSE
       \STATE Store a subset of data from $\mathcal{T}_t$ in $\mathcal{E}$
    \ENDIF
    
    \STATE Estimate $\mathcal{W}_{jj}^{{(t)}}$ for task $\mathcal{T}_t$ as per Eq.~\ref{eq:fisher-sc-exp}
    \STATE Snapshot parameters $\Phi_{t}^*$
    \STATE Append $(\Phi_{t}^*, \mathcal{W}_{jj}^{(t)})$ to EWC record list $\mathcal{R}$
\ENDFOR

\STATE \textbf{Return} $q_{\phi}(\theta \mid x)$

\end{algorithmic}
\end{algorithm}

\begin{algorithm}[h]
\caption{\textsc{KMedoidUpdate}}
\label{alg:k-medoid}
\begin{algorithmic}[1]
\STATE \textbf{Input:} Current memory $\mathcal{E}$, new task $x_t^*$, posterior samples $\{\theta_t^{(s)}\}$, memory size $K$
\STATE \textbf{Output:} Updated memory $\mathcal{E}$

\STATE Add $(x_t^*, \{\theta_t^{(s)}\})$ to $\mathcal{E}$

\IF{$|\mathcal{E}| \le K$}
    \STATE \textbf{return} $\mathcal{E}$
\ELSE
    \STATE Form posterior representations $\{\bar{\theta}_i\}$
    \STATE Apply K-medoids clustering in parameter space and let $\mathcal{M}$ be the set of indices of cluster representatives
    \STATE Retain only cluster representatives and discard the rest 
    ($\mathcal{E} \leftarrow \{(x_i^*, \{\theta_i^{(s)}\}) : i \in \mathcal{M}\}$)
    \STATE \textbf{return} $\mathcal{E}$
\ENDIF
\end{algorithmic}
\end{algorithm}

\section{Experiment 1: Dataset and model description} \label{sec:lr_dataset}

For continual learning, we utilize nine real-world regression datasets, each treated as a sequential task in the CL setting (Tasks 1--9). For each dataset, we select five representative predictors and consider a single continuous response variable. The datasets are obtained from standard public repositories, including OpenML\citep{JMLR:v22:19-920}, scikit-learn \citep{pedregosa2011scikit}, and the UCI Machine Learning Repository \citep{lichman2013uci}.

\paragraph{Task 1: Boston Housing.}
We use the Boston Housing dataset \citep{harrison1978hedonic} from OpenML (version 1). 
Predictors: CRIM (crime rate), ZN (residential land proportion), RM (average number of rooms), AGE (proportion of old units), and DIS (distance to employment centers). 
Response: Median house value.

\paragraph{Task 2: Auto MPG.}
We use the Auto MPG dataset from OpenML (version 1). 
Predictors: cylinders, displacement, horsepower, weight, and acceleration. 
Response: Miles per gallon (mpg), a continuous measure of fuel efficiency.

\paragraph{Task 3: California Housing.}
We use the California Housing dataset \citep{pace1997sparse} from scikit-learn. 
Predictors: MedInc (median income), HouseAge, AveRooms, AveOccup, and Population. 
Response: Median house value.

\paragraph{Task 4: Diabetes.}
We use the Diabetes dataset from scikit-learn. 
Predictors: bmi (body mass index), bp (blood pressure), s1 and s5 (serum measurements), and age. 
Response: Quantitative measure of disease progression one year after baseline.

\paragraph{Task 5: Energy Efficiency.}
We use the Energy Efficiency dataset (UCI ID 242). 
Predictors: Relative Compactness (X1), Surface Area (X3), Wall Area (X6), Overall Height (X7), and Glazing Area (X8). 
Response: Heating Load (Y1).

\paragraph{Task 6: Bike Sharing.}
We use the Bike Sharing dataset (UCI ID 275). 
Predictors: mnth (month), temp (temperature), hum (humidity), windspeed, and weekday. 
Response: Bike rental count.

\paragraph{Task 7: Concrete Compressive Strength.}
We use the Concrete Compressive Strength dataset (UCI ID 165). 
Predictors: Cement, Water, Superplasticizer, Coarse Aggregate, and Fly Ash. 
Response: Concrete compressive strength.

\paragraph{Task 8: Liver Disorders.}
We use the Liver Disorders dataset (UCI ID 60). 
Predictors: mcv, alkphos, sgpt, sgot, and gammagt (blood test measurements). 
Response: Continuous liver disorder indicator.

\paragraph{Task 9: Superconductivity.}
We use the Superconductivity dataset (UCI ID 464) \citep{hamidieh2018data}. 
Predictors: mean\_atomic\_mass, mean\_Valence, mean\_atomic\_radius, mean\_ElectronAffinity, and mean\_FusionHeat. 
Response: Critical temperature of superconducting materials.

Across all tasks, input features are standardized prior to training. The dataset for Task 0 was simulated from the prior predictive distribution. For each task, we partition the data into 10 subsets with 100 observations each. The observations were sampled independently from the full dataset with replacement.

For the summary network, we use set-transformer \citep{lee2019set} an attention-based permutation-invariant neural network to learn 50-dimensional embeddings that are maximally informative for posterior inference. For the neural posterior estimator $q(\theta \mid x)$, we use a neural spline flow \citep{durkan2019neural} with 6 coupling layers of 128 units each utilizing exponential linear unit activation functions and a multivariate unit Gaussian latent space. These settings were the same for both the standard simulation-based pre-training and for the training of unsupervised CL methods.

The posterior network and summary networks are jointly trained. For SB baseline, we use online training with 64 simulation batches and a batch size of 32 for 50 epochs with a learning rate of $10^{-3}$. For subsequent unsupervised training of naive SC, SC-ER, SC-EWC, SC-ER-EWC, and test-time SC, we use the Adam optimizer with a cosine decay learning rate schedule. The initial learning rate is initialized as $10^{-3}.t$, where $t$ is the task index, and is annealed following a cosine schedule over all training steps. Training is done for 30 epochs for each task with each step utilizing the entire data set for computing the SC loss, with the variance target computed over 16 samples from the current posterior approximation \citep{mishra2025robust}.

\section{Experiment 1: Additional Results} \label{sec:app_res1}

\begin{figure}[h]
    \centering
    \includegraphics[width=\linewidth]{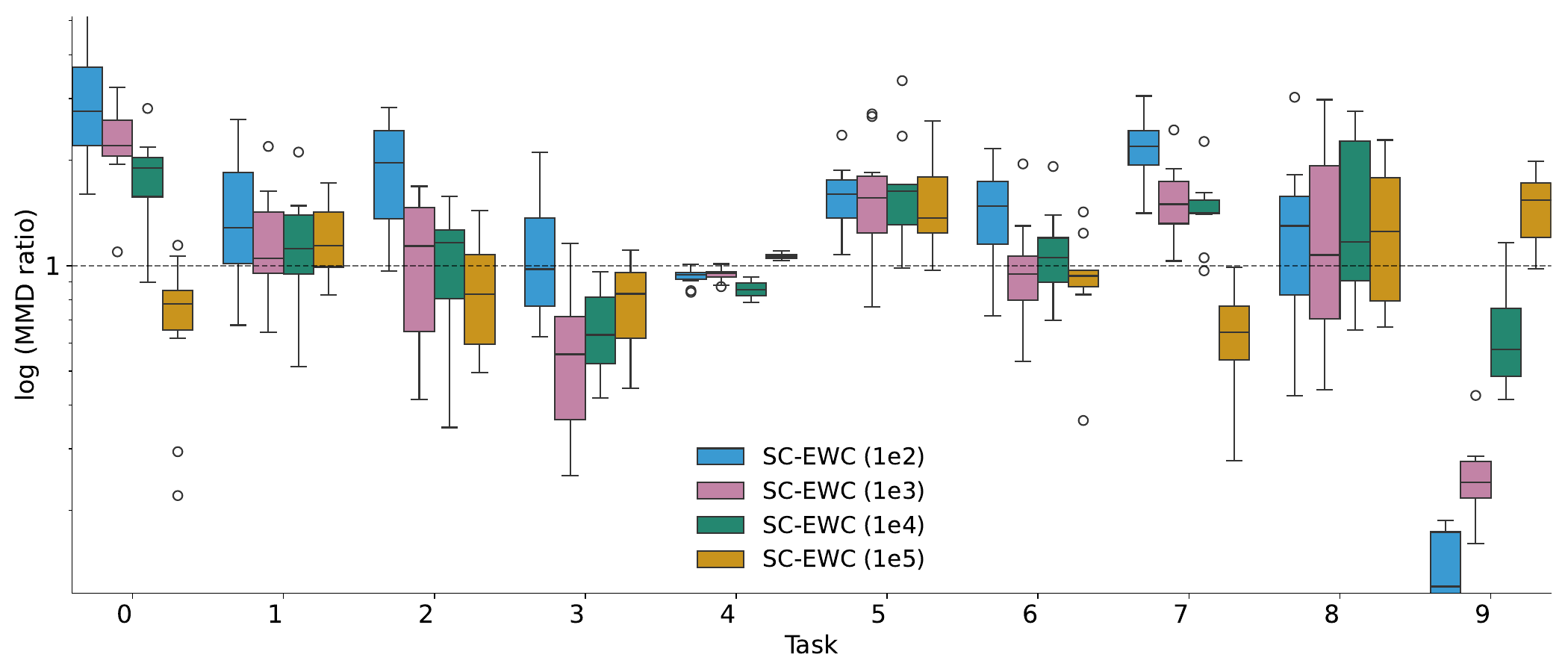}
    \caption{\textbf{Experiment 1.} MMD ratio in log scale across CL tasks (0–9) for SC-EWC with different values of EWC hyperparameter $\lambda$. Boxes summarize variability across subsets. Dashed line marks parity with simulation-based (SB) baseline (ratio = 1). $\lambda = 10^{2}$ shows forgetting while other variants perform similarly.}
    \label{fig:mmd_ewc_lr}
\end{figure}

\begin{figure}[h!]
    \centering
    \includegraphics[width=\linewidth]{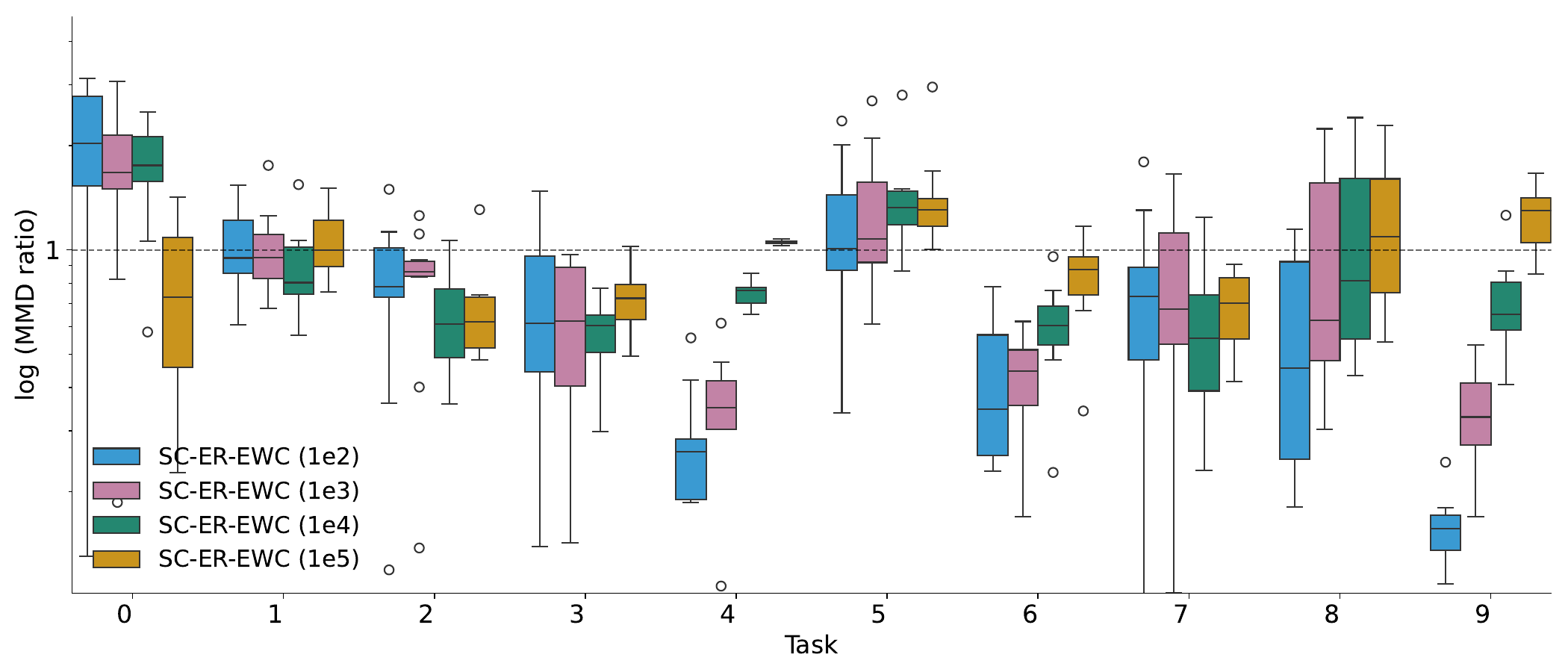}
    \caption{\textbf{Experiment 1.} MMD ratio in log scale across CL tasks (0–9) for SC-ER-EWC with different values of EWC hyperparameter $\lambda$. Boxes summarize variability across subsets. Dashed line marks parity with simulation-based (SB) baseline (ratio = 1). The performance remains largely stable even for smaller $\lambda$ values, indicating that the robustness improvements are primarily attributable to episodic replay. The addition of EWC provides only marginal benefits in this setting.}
    \label{fig:mmd_er_ewc_lr}
\end{figure}

We conducted a sensitivity analysis to evaluate the effect of the EWC regularization strength parameter, $\lambda$, which controls the degree of constraint imposed on previously learned parameters. Specifically, we trained both SC-EWC and SC-ER-EWC with $\lambda$ values ranging from $10^{2}$ to $10^{5}$, while keeping all other hyperparameters fixed. The corresponding MMD ratio values across tasks are reported in Figures~\ref{fig:mmd_ewc_lr} and~\ref{fig:mmd_er_ewc_lr} for SC-EWC and SC-ER-EWC, respectively. 

For SC-EWC, we observe that the variant with $\lambda = 10^{2}$ exhibits noticeably stronger catastrophic forgetting on several tasks compared to configurations with larger $\lambda$ values. Increasing the regularization strength mitigates this effect, and overall performance remains relatively stable for $\lambda \in {10^{3}, 10^{4}, 10^{5}}$, with $\lambda = 10^{3}$ yielding the best performance across tasks. 

In contrast, for SC-ER-EWC, performance remains largely stable even for smaller $\lambda$ values, indicating that the robustness improvements are primarily attributable to episodic replay. The addition of EWC provides only marginal benefits in this setting. However, when $\lambda$ is increased to $10^{5}$, performance deteriorates relative to other configurations, suggesting that overly strong regularization can hinder adaptation and negatively impact robustness.

\begin{figure}[h!]
    \centering
    \includegraphics[width=\linewidth]{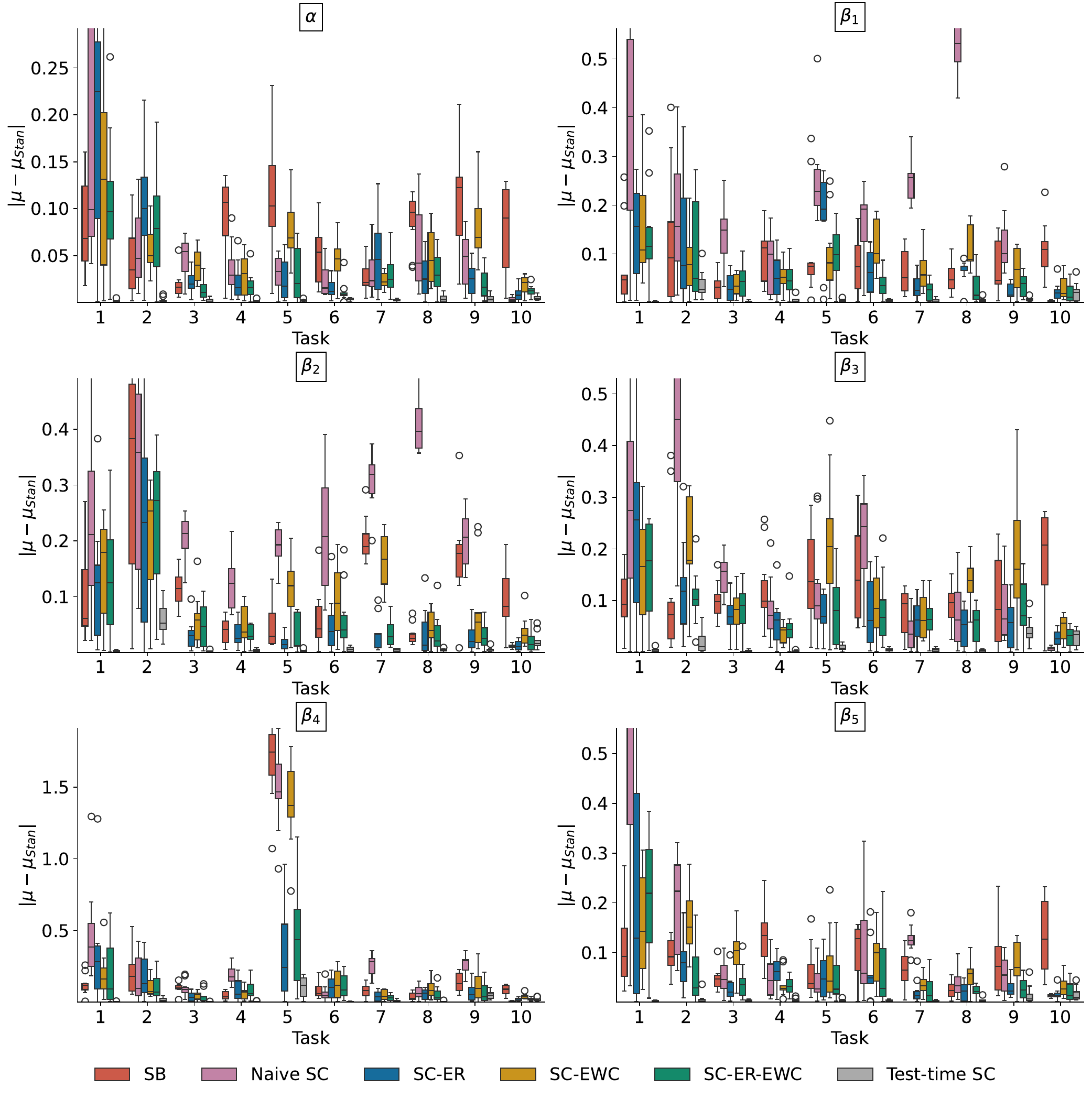}
    \caption{\textbf{Experiment 1.} Absolute mean bias of linear regression parameters compared to Stan based estimates across CL tasks (0–9) for various methods. Boxes summarize variability across subsets. SB and naive SC show catastrophic forgetting while CL methods perform consistently better across tasks.}
    \label{fig:mean_bias_lr}
\end{figure}

\begin{figure}[h!]
    \centering
    \includegraphics[width=\linewidth]{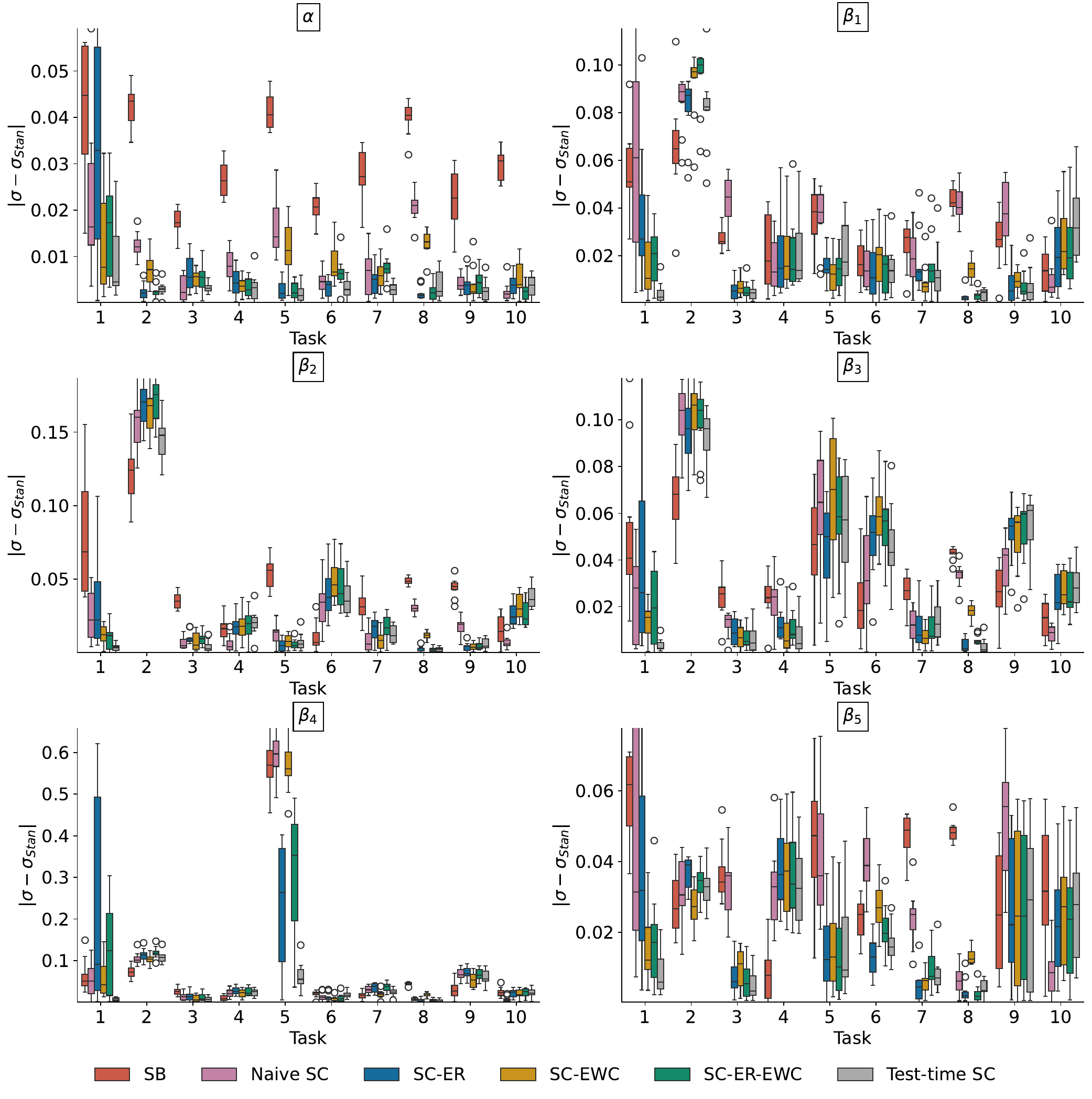}
    \caption{\textbf{Experiment 1.} Absolute standard deviation bias of linear regression parameters compared to Stan based estimates across CL tasks (0–9) for various methods. Boxes summarize variability across subsets.}
    \label{fig:std_bias_lr}
\end{figure}

\begin{figure}[h!]
    \centering
    \includegraphics[width=\linewidth]{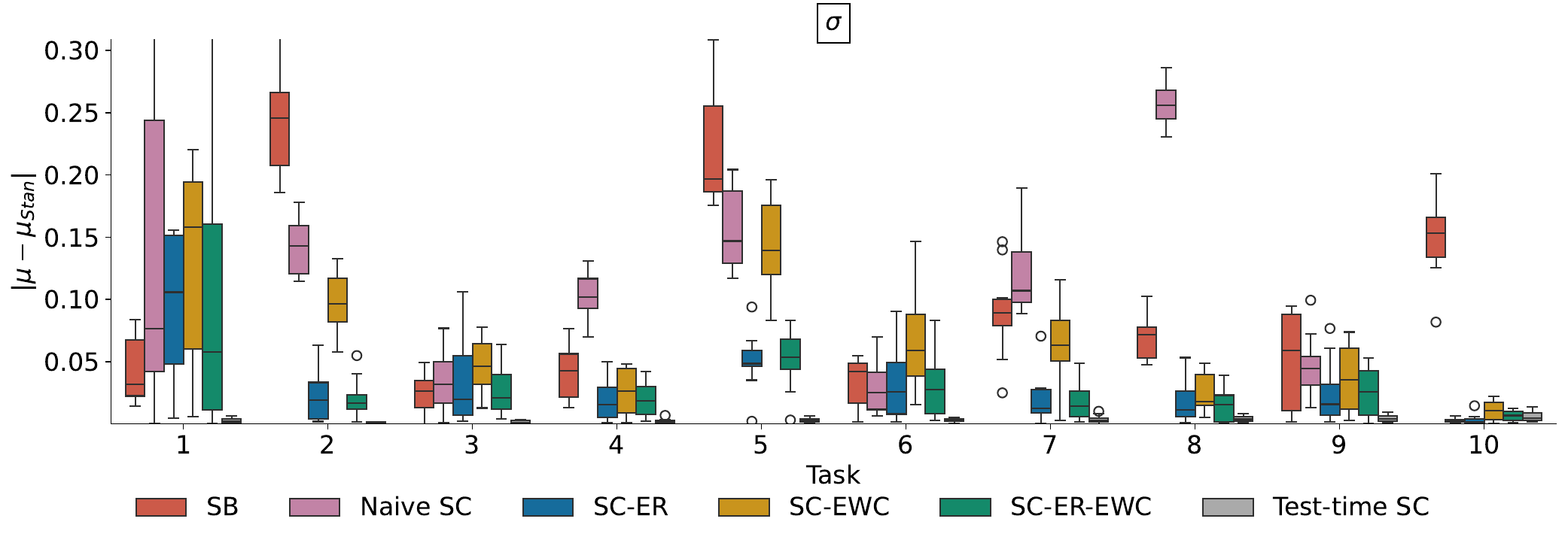}
    \caption{\textbf{Experiment 1.} Absolute mean bias of linear regression parameter $\sigma$ compared to Stan based estimates across CL tasks (0–9) for various methods. Boxes summarize variability across subsets. SB and naive SC show high bias across tasks while other methods perform consistently better.}
    \label{fig:mean_sigma_lr}
\end{figure}

\begin{figure}[h!]
    \centering
    \includegraphics[width=\linewidth]{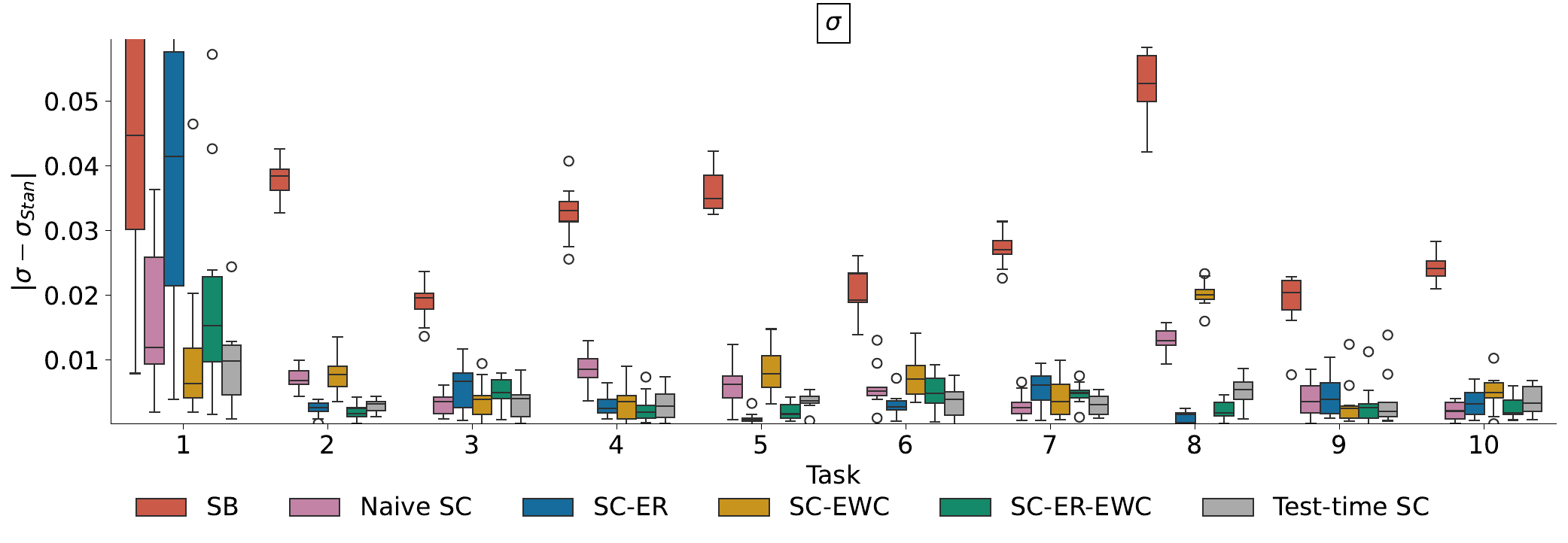}
    \caption{\textbf{Experiment 1.} Absolute standard deviation bias of linear regression parameter $\sigma$ compared to Stan based estimates across CL tasks (0–9) for various methods. Boxes summarize variability across subsets. SB shows high bias across tasks while other methods perform consistently better.}
    \label{fig:sigma_std_lr}
\end{figure}

\FloatBarrier
\section{Experiment 2: Dataset and Model Description} \label{sec:app_air_model}

The parameters $\alpha_j$ denote country-level intercepts, $\beta_j$ are the autoregressive coefficients, $\gamma_j$ are the regression coefficients of household debt and $\delta_j$ are the regression coefficients of GDP per capita, and $\sigma_j$ is the standard deviation of the noise term. This model has previously been applied within ABI in \cite{habermann2024amortized, mishra2025robust, kucharsky2025towards}. Following standard practice for autoregressive models, we regress on time period differences to mitigate non-stationarity. This is critical for simulation-based inference because $\beta_j > 1$ can induce exponential growth and quickly produce unrealistic air traffic volumes. Moreover, amortizing over covariate spaces, such as varying GDP per capita between countries, can lead to model misspecification if such fluctuations are underrepresented in training. 

For the air traffic model defined in Section \ref{sec:air}, we set independent priors on the parameters as follows:

\begin{align*}
    \alpha_j &\sim \mathcal{N}(0, 0.5) \hspace{1.5cm} \beta_j \sim \mathcal{N}(0, 0.2) \\ 
    \gamma_j &\sim \mathcal{N}(0, 0.5) \hspace{1.5cm} \delta_j \sim \mathcal{N}(0, 0.5) \\
    \mathrm{log}(\sigma_j) & \sim \mathcal{N}(-1, 0.5). 
\end{align*}

For the summary network, we use a long short-term memory layer with 64 output dimensions followed by two dense layers with output dimensions of 256 and 64. For the neural posterior estimator $q(\theta \mid x)$, we use a neural spline flow \citep{durkan2019neural} with 6 coupling layers of 128 units each utilizing exponential linear unit activation functions and a multivariate unit Gaussian latent space. These settings were the same for both the standard simulation-based pre-training and for the training of unsupervised CL methods. 

 The posterior network and summary networks are jointly trained. For SB baseline, we use online training with 64 simulation batches and a batch size of 32 for 50 epochs with a learning rate of $10^{-3}$. For subsequent unsupervised training of naive SC, SC-ER, SC-EWC, SC-ER-EWC, and test-time SC, we use the Adam optimizer with a cosine decay learning rate schedule. The initial learning rate is initialized as $10^{-3}.t$, where $t$ is the task index, and is annealed following a cosine schedule over all training steps. Training is done for 30 epochs for each task with each step utilizing the entire data set for computing the SC loss, with the variance target computed over 16 samples from the current posterior approximation \citep{mishra2025robust}.

\FloatBarrier
\section{Experiment 2: Additional Results} \label{sec:air_results}

\begin{figure}[h]
    \centering
    \includegraphics[width=\linewidth]{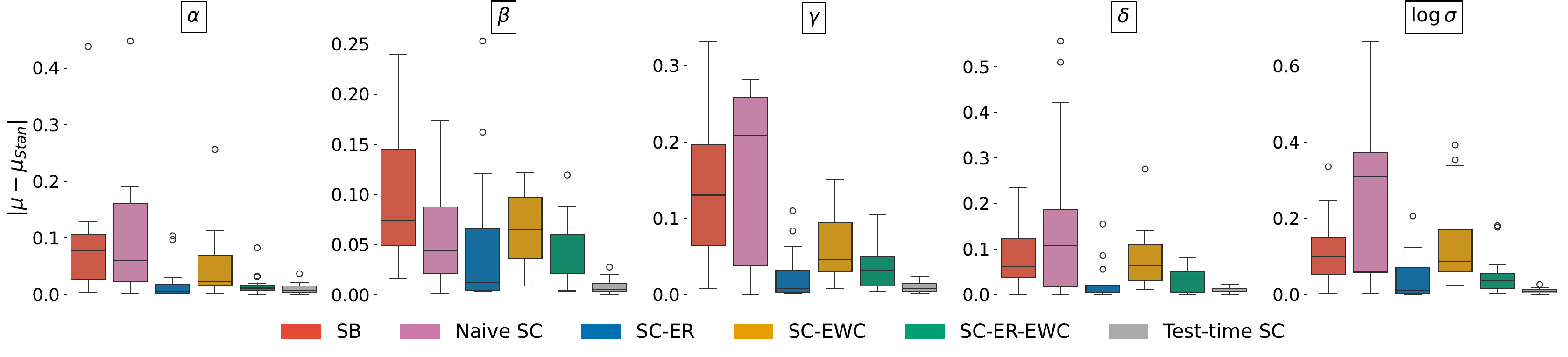}
    \caption{\textbf{Experiment 2.} Absolute mean bias of air traffic model parameters compared to Stan based estimates for various methods aggregated over 15 CL tasks. Boxes summarize variability of parameters across tasks. SB and naive SC show high bias across tasks while other methods perform consistently better.}
    \label{fig:mean_bias_air}
\end{figure}

\begin{figure}[h]
    \centering
    \includegraphics[width=\linewidth]{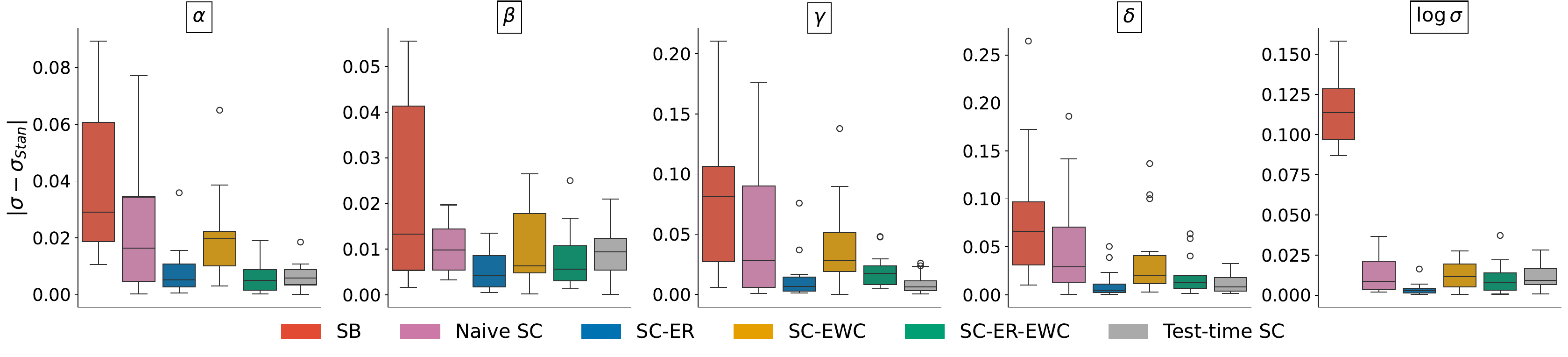}
    \caption{\textbf{Experiment 2.} Absolute standard deviation bias of air traffic model parameters compared to Stan based estimates for various methods aggregated over 15 CL tasks. Boxes summarize variability of parameters across tasks. SB and naive SC show high bias across tasks while other methods perform consistently better.}
    \label{fig:std_bias_air}
\end{figure}

\begin{figure}[h]
    \centering
    \includegraphics[width=\linewidth]{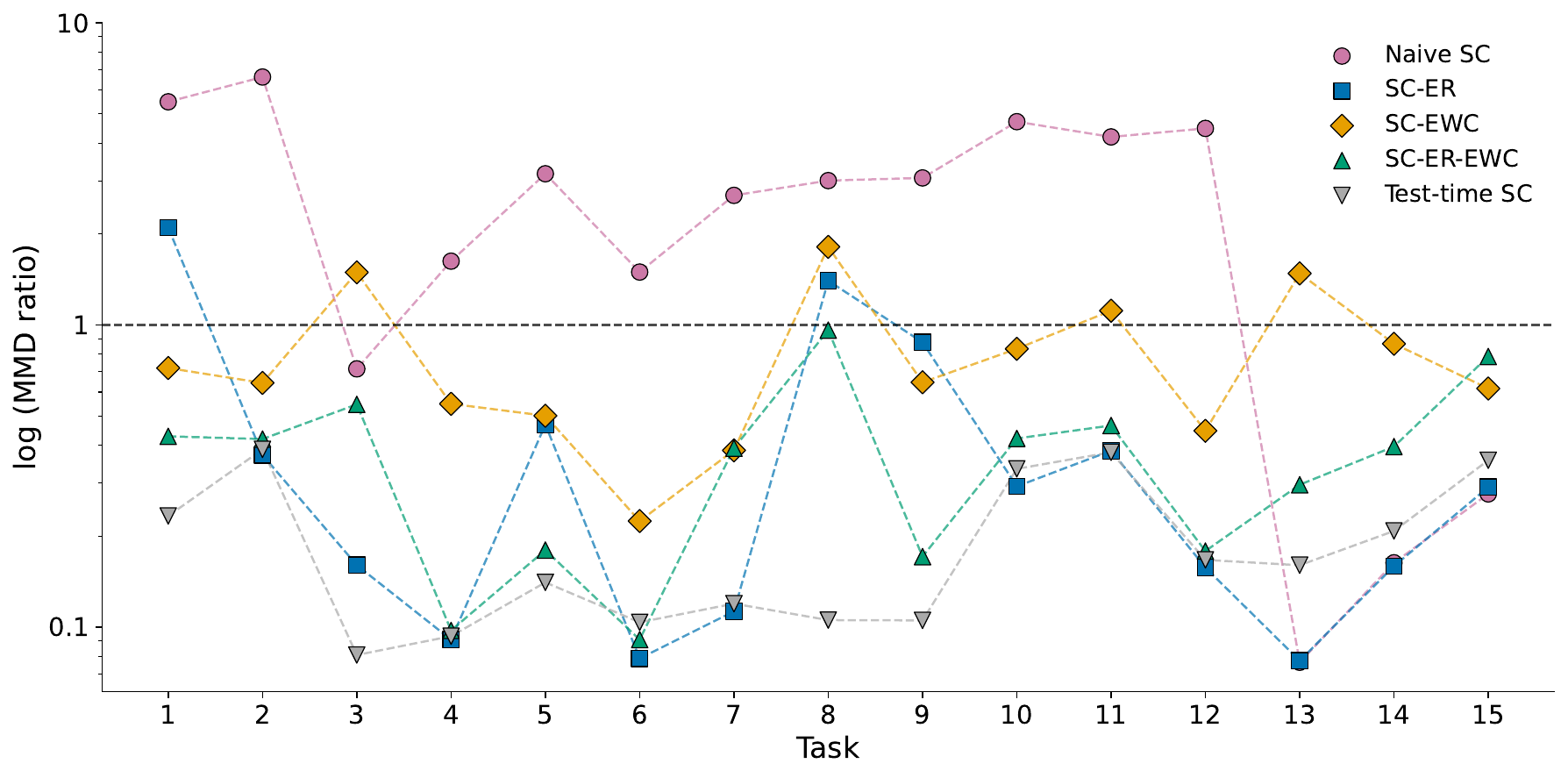}
    \caption{\textbf{Experiment 2.} MMD ratio in log scale across CL tasks (1–15) for different methods. Dashed line marks parity with simulation-based (SB) baseline (ratio = 1). Naive SC shows catastrophic forgetting in CL setting whereas our proposed methods mitigate forgetting and provide better posterior estimates compared to SB and Naive SC. Test-time SC also gives accurate posterior estimates.}
    \label{fig:air_mmd_full}
\end{figure}

\begin{figure}[h]
    \centering
    \includegraphics[width=0.7\linewidth]{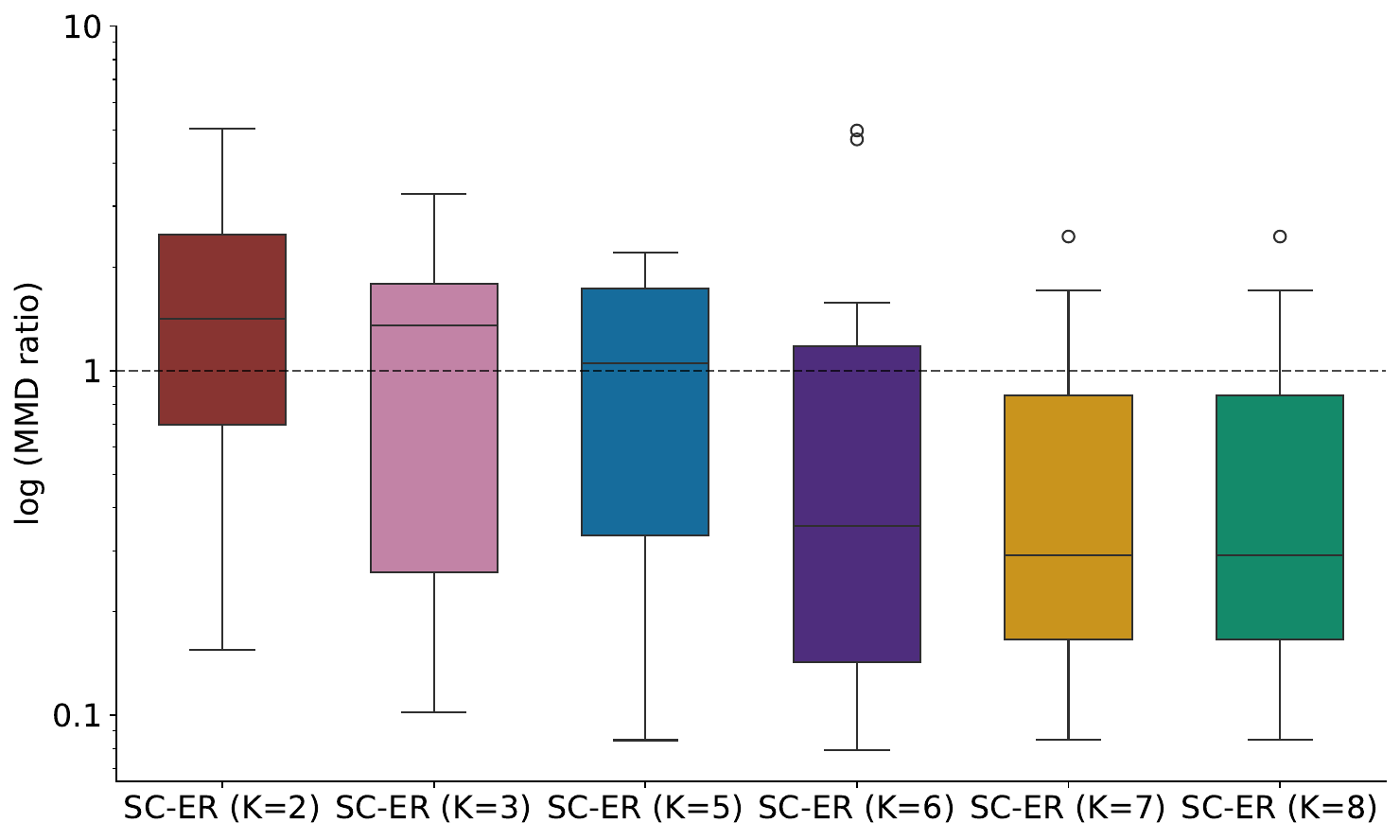}
    \caption{\textbf{Experiment 2.} MMD ratio in log scale aggregated over fifteen CL tasks for SC-ER with varying replay memory buffer $K$.  Dashed line marks parity with simulation-based baseline (ratio = 1). For values of $K \leq 6$ the performance of SC-ER starts to decrease whereas for $K \geq 7$, good performance is observed.}
    \label{fig:k-med}
\end{figure}

\begin{figure}[h]
    \centering
    \includegraphics[width=0.7\linewidth]{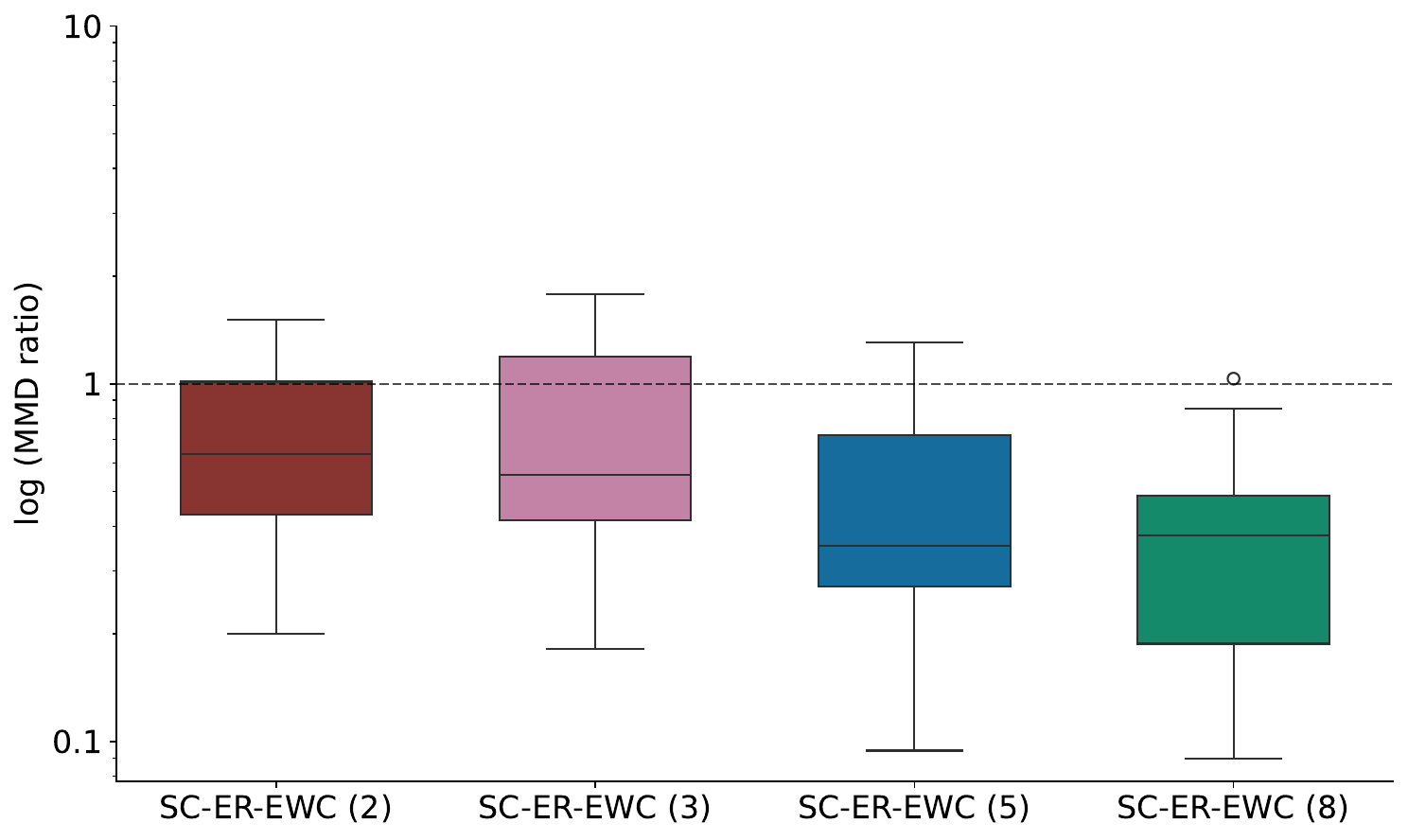}
    \caption{\textbf{Experiment 2.} MMD ratio in log scale aggregated over fifteen CL tasks for SC-ER-EWC with varying replay memory buffer $K$.  Dashed line marks parity with simulation-based baseline (ratio = 1). For values of $K = 2,3$ the performance of SC-ER-EWC decreases but still remain better than SB baseline as well as SC-ER for same K-value. It signifies that addition of EWC leads to enhanced performance when replay memory is limited.}
    \label{fig:k-med-er-ewc}
\end{figure}

\FloatBarrier
\section{Experiment 3: Dataset and Model Description} \label{sec:app_diffusion}

The eight empirical datasets originally come from studies done by \citet{chetverikov2017blame}, \citet{hedge2018reliability}, and \citet{pratte2010exploring} and were extracted from the database provided by \citet{haaf2025attentional} under the dataset ID numbers: 1, 5, 6, 7, 8, 38, 39, and 40. These datasets contain responses of participants on the Stroop task \citep{stroop1935studies}, Flanker task, \citep{eriksen1974effects}, or Simon task \citep{simon1990effects}. The number of participants in each data set ranges between 30 and 60. For convenience, we recoded the response time and accuracy data so that incorrect responses are coded as negative response times. Further, we also retained 180 trials per participant (90 congruent and 90 incongruent trials), so that the networks are trained on data with constant sample size.

The racing diffusion model \citep{tillman2020sequential} posits that each response alternative (correct vs.\ incorrect) is represented by its own noisy evidence accumulator. Each accumulator starts at zero and follows a Wiener process with drift (where $W_t$ denotes a standard Wiener process),
\begin{equation}
X_t = \nu~\mathrm{d}t + \sigma~\mathrm{d}W_t,
\qquad X_0 = 0.
\end{equation}
A response is generated when one of the accumulators first reaches the decision boundary $\alpha$. The observed response time corresponds to the hitting time of the winning accumulator plus a non-decision time component $t_0$. Consequently, the model yields response times ($rt$, in seconds) together with accuracies (correct/incorrect) on each trial. For training, incorrect responses were encoded as negative values of the associated $rt$, allowing the networks to condition on a single variable that jointly represents speed and accuracy.

We used the following priors on the parameters,
\begin{equation}
\begin{aligned}
\log \alpha \sim \mathcal{N}(0, 0.5) \
\log \nu_{\text{correct}} &\sim \mathcal{N}(0, 0.5), \
\log \nu_{\text{incorrect}} &\sim \mathcal{N}(0, 0.5), \
\mathrm{logit}~\tau &\sim \mathcal{N}(0, 1),
\end{aligned}
\end{equation}
with $\sigma = 1$. The parameter $\tau$ determines the non-decision time via
\begin{equation}
t_0 = \min(rt)\frac{\tau}{1-\tau}.
\end{equation}
This reparameterization allows estimation of $\mathrm{logit}~\tau$ on an unconstrained scale while ensuring $t_0 \in (0, \min(rt))$.

Both the neural posterior and neural likelihood estimators are implemented as normalizing flows \citep{dinh2016density} with six spline coupling layers. Each coupling layer contains two dense layers with 256 hidden units and mish activations \citep{misra2019mish}. The summary network follows a DeepSet architecture \citep{zaheer2017deep} comprising two equivariant modules (each with two dense layers and 64 hidden units), an invariant module with two dense layers and 32 hidden units in both the inner and outer components, an additional invariant module with two dense layers of 32 hidden units, and an output layer with 30 units. All layers except the output layer use SiLU activations.

SB training followed an online regime (100 epochs, 64 steps per epoch with batch size of 64). SC training consisted of 200 training steps, with each step utilizing the entire data set for computing the SC loss, with the variance target computed over 128 samples from the current posterior approximation \citep{mishra2025robust}. Because the SC loss was unstable, we restored weights of the networks that gave the smallest SC loss during the 200 training steps, to ensure that the results are not affected by a random fluctuation at the final step.

\FloatBarrier
\section{Experiment 3: Additional Results} \label{sec:app_diffusion_results}

\begin{figure*}[h]
    \centering
    \includegraphics[width=0.99\linewidth]{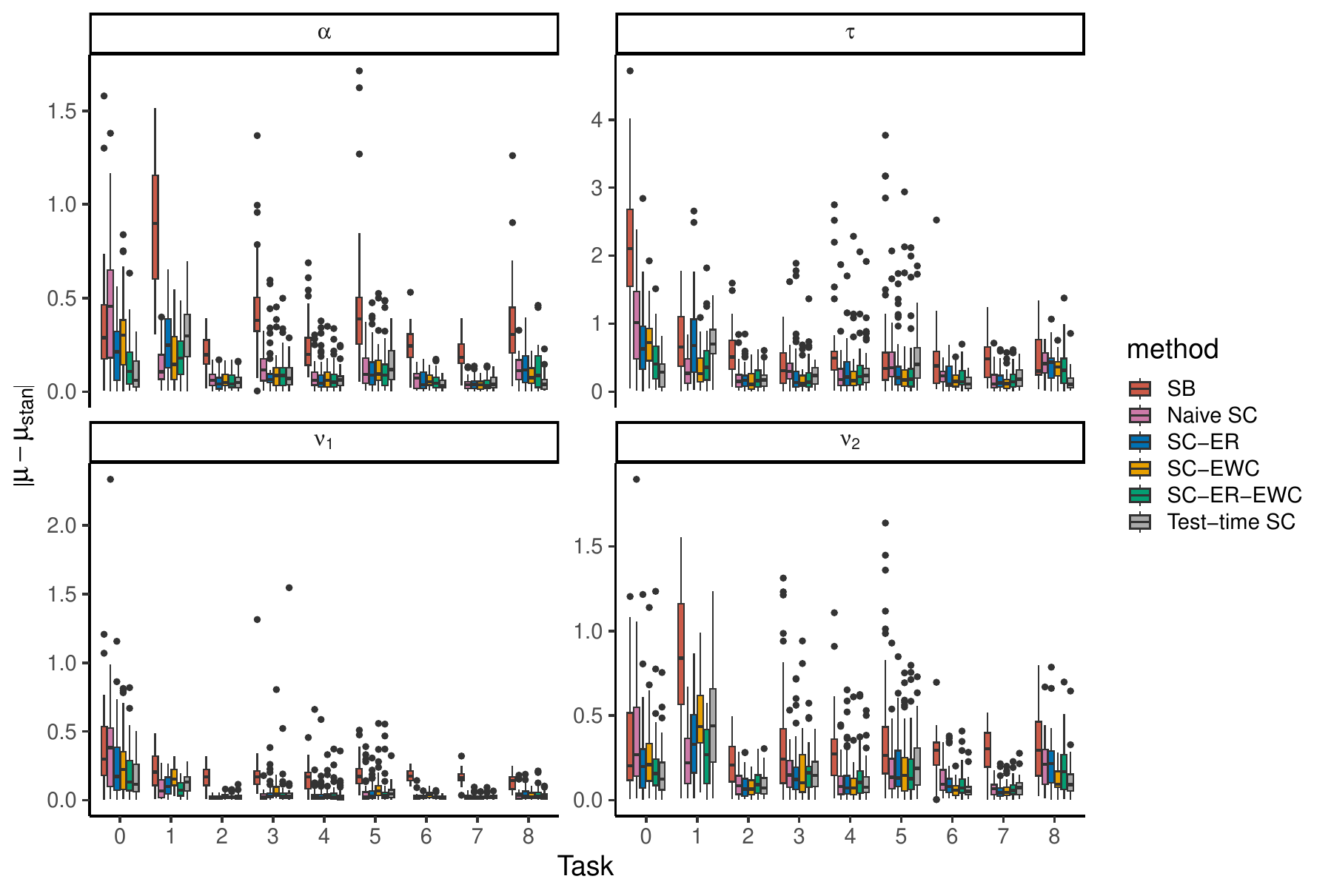}
    \caption{\textbf{Experiment 3.} Absolute bias of the posterior mean of the marginal posteriors of the racing diffusion model.}
    \label{fig:diffusion_mean_error}
\end{figure*}

\begin{figure*}[h]
    \centering
    \includegraphics[width=0.99\linewidth]{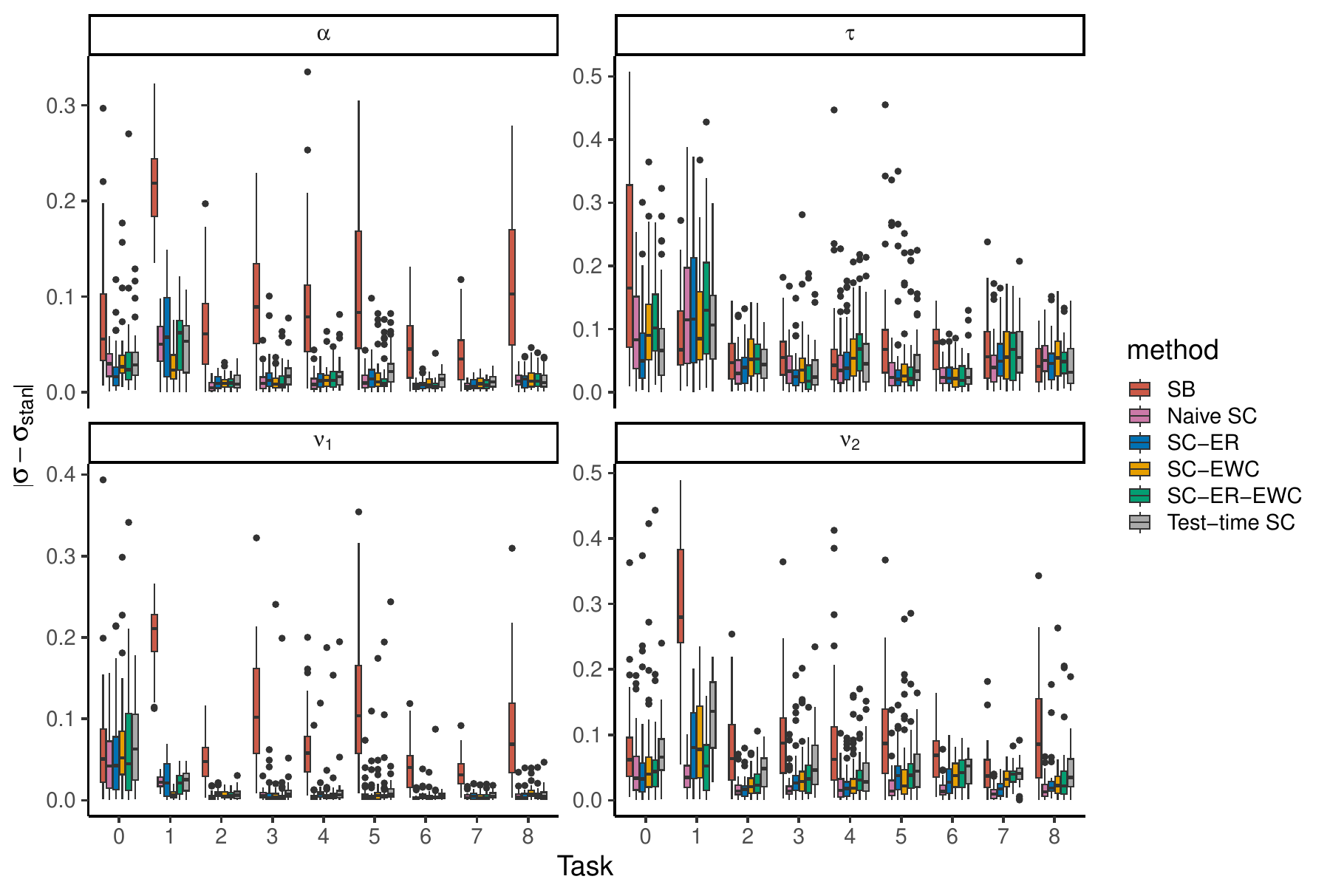}
    \caption{\textbf{Experiment 3.} Absolute bias of the posterior standard deviation of the marginal posteriors of the racing diffusion model.}
    \label{fig:diffusion_sd_error}
\end{figure*}

\end{document}